\documentclass{article} 
\usepackage[final]{colm2026_conference}

\usepackage{microtype}
\usepackage{hyperref}
\usepackage{url}
\usepackage{booktabs}

\usepackage{amsmath,amsfonts,bm}

\def\eqref#1{(\ref{#1})}

\def\1{\bm{1}}

\DeclareMathAlphabet{\mathsfit}{\encodingdefault}{\sfdefault}{m}{sl}
\SetMathAlphabet{\mathsfit}{bold}{\encodingdefault}{\sfdefault}{bx}{n}

\newcommand{\btheta}{{\boldsymbol{\theta}}}

\usepackage{wrapfig}

\usepackage{algorithm}
\usepackage{algpseudocode}

\usepackage{enumitem} 

\usepackage{multirow,mathtools }

\usepackage{pifont}
\usepackage{color, colortbl}

\usepackage{blindtext}
\usepackage{lipsum}

\usepackage{multirow}
\usepackage{listings}

\usepackage{bbm}

\usepackage [english]{babel}
\usepackage[autostyle, english = american]{csquotes}

\usepackage{pifont}
\usepackage{url}
\usepackage[most]{tcolorbox}

\usepackage{lipsum}
\usepackage{soul}
\usepackage{xcolor}
\usepackage{wrapfig}
\usepackage{multirow,mathtools } 

\usepackage{adjustbox}
\MakeOuterQuote{"}

\usepackage{microtype}
\usepackage{graphicx}
\usepackage{booktabs} 

\usepackage{hyperref}

\usepackage{amsmath}

\usepackage[capitalize,noabbrev]{cleveref}

\usepackage{nicefrac}       

\usepackage{tablefootnote}

\usepackage{float} 
\usepackage{diagbox}

\newcommand{\Def}[0]{\mathrel{\mathop:}=}

\usepackage{pifont}
\newcommand{\cmark}{\textcolor{green!50!black}{\ding{51}}} 
\newcommand{\qmark}{\textcolor{red!70!black}{\textbf{?}}}  

\usepackage{color, colortbl}
\definecolor{Gray}{gray}{0.93}
\definecolor{Orange}{rgb}{1,0.5,0}
\definecolor{DGray}{gray}{0.83}
\definecolor{LightCyan}{rgb}{0.88,1,1}

\definecolor{WarnREd}{rgb}{1,0.4,0.4}
\definecolor{WarnOrange}{rgb}{1,0.682,0.502}
\definecolor{WarnPink}{rgb}{0.9176, 0.7215, 0.7215}
\definecolor{GoodGreen}{rgb}{0.5019, 0.9215, 0.6039}

\usepackage[T1]{fontenc}

\definecolor{styleblue}{HTML}{504099}
\definecolor{mypurple}{HTML}{9391ff}

\definecolor{bluegray}{rgb}{0.4, 0.6, 0.8}
\definecolor{ceruleanblue}{rgb}{0.16, 0.32, 0.75}

\hypersetup{
colorlinks=true,
citecolor=ceruleanblue,
linkcolor=ceruleanblue,
urlcolor=black
}

\definecolor{LightCyan}{rgb}{0.88,1,1}

\usepackage{lineno}
\usepackage{fontawesome5}

\definecolor{darkblue}{rgb}{0, 0, 0.5}
\hypersetup{colorlinks=true, citecolor=darkblue, linkcolor=darkblue, urlcolor=darkblue}

\title{
Who Built This Model? Tracing LLM Lineage via Spectral Fingerprints in Weight Space
}

\author{
\textbf{Yiwei Chen}$^{\dag,\star}$
\quad
\textbf{Bingqi Shang}$^{\dag,\star}$
\quad
\textbf{Sijia Liu}$^{\dag}$
\\[0.3em]
$^\dag$Michigan State University
\quad
$^\star$Equal contribution
\\[0.6em]
\href{https://github.com/OPTML-Group/LLM-Biometrics}
{\faGithub\ GitHub}
\qquad
\href{https://yiwei-chenn.github.io/llm-biometrics}
{\faGlobe\ Project}
}

\definecolor{myblue}{RGB}{70,130,180}
\definecolor{myorange}{RGB}{255,140,0}

\everydisplay{\small}

\begin{document}

\ifcolmsubmission
\linenumbers
\fi

\maketitle

\begin{abstract}
Open-weight large language models (LLMs) are increasingly developed through complex, multi-stage pipelines, leading to intricate lineage relationships that reflect model origin, ownership, and evolution. 
Understanding these relationships is important for model provenance, governance, and supply-chain integrity.
In this work, we investigate the notion of LLM ``biometrics'' (analogous to human biometrics) to ask whether LLMs exhibit intrinsic fingerprints in weight space alone, without access to input data, that reveal their origin and lineage. 
We formulate this as a lineage discrimination problem, distinguishing among independent-origin, same-series, and shared-base models.
To characterize these relationships, we propose a unified \textit{geometric fingerprinting} framework that analyzes weight matrices from two complementary perspectives: 
(i) spectral energy, captured by singular value distributions to encode global magnitude patterns, and 
(ii) subspace alignment, quantified via subspace deviations to capture directional geometry.
Our analysis uncovers a clear hierarchy of structural similarity in weight space: spectral energy reliably distinguishes independently trained models and different model families, while subspace alignment enables fine-grained discrimination among closely related models, including variations in dataset scale and post-training procedures.
Extensive experiments on over 110 diverse open-weight LLM pairs demonstrate that weight-space geometry provides a robust and interpretable signal for model lineage, enabling coarse-grained regime separation and fine-grained discrimination within shared-base models.
\end{abstract}
\section{Introduction}
\label{sec:intro}

The rapid growth of open-weight large language models (LLMs) has created a complex ecosystem shaped by multi-stage pipelines spanning pretraining, fine-tuning, and alignment~\citep{touvron2023llama, ouyang2022training, bai2022constitutional, qwen2.5, team2024gemma}.
This has led to a proliferation of derived models from shared base models across organizations, scales, and post-training procedures~\citep{chung2024scaling, wang2023self, zhou2023lima, yang2025qwen3,chen2026unlearning,shang2025forgetting}, forming intricate lineage relationships among models.
Understanding these relationships, namely \emph{how they are related, and how they evolve}, is critical for model governance and supply-chain integrity \citep{carlini2021extracting,pahune2025importance}.
This motivates the notion of \emph{LLM biometrics}, which aims to identify intrinsic signatures of model identity and lineage.

Despite this need, identifying relationships between LLMs remains highly challenging.
Existing approaches primarily rely on \emph{black-box} behavioral fingerprints, such as prompt-response probing, watermarking, or output distribution analysis~\citep{wu2025llmdna, kuditipudi2025blackbox, nikolic2025model, tsai2025rofl}.
These methods rely on input data, are sensitive to prompt design, and provide limited insight into intrinsic ``fingerprints'' or ``biometrics'' derived solely from LLM weights.
Recent \emph{white-box} approaches instead analyze model parameters directly, for example, using weight statistics or  similarity~\citep{zhang2024reef, zeng2025awm, zeng2024huref, yoon2025intrinsic}.
However, these methods typically reduce complex weight-space information to coarse summary statistics. In this paper, we will show that such approaches can fail to distinguish models in many lineage discrimination scenarios.

\begin{figure}[t]
\vspace{-0mm}
\centering
\includegraphics[width=0.95\linewidth]{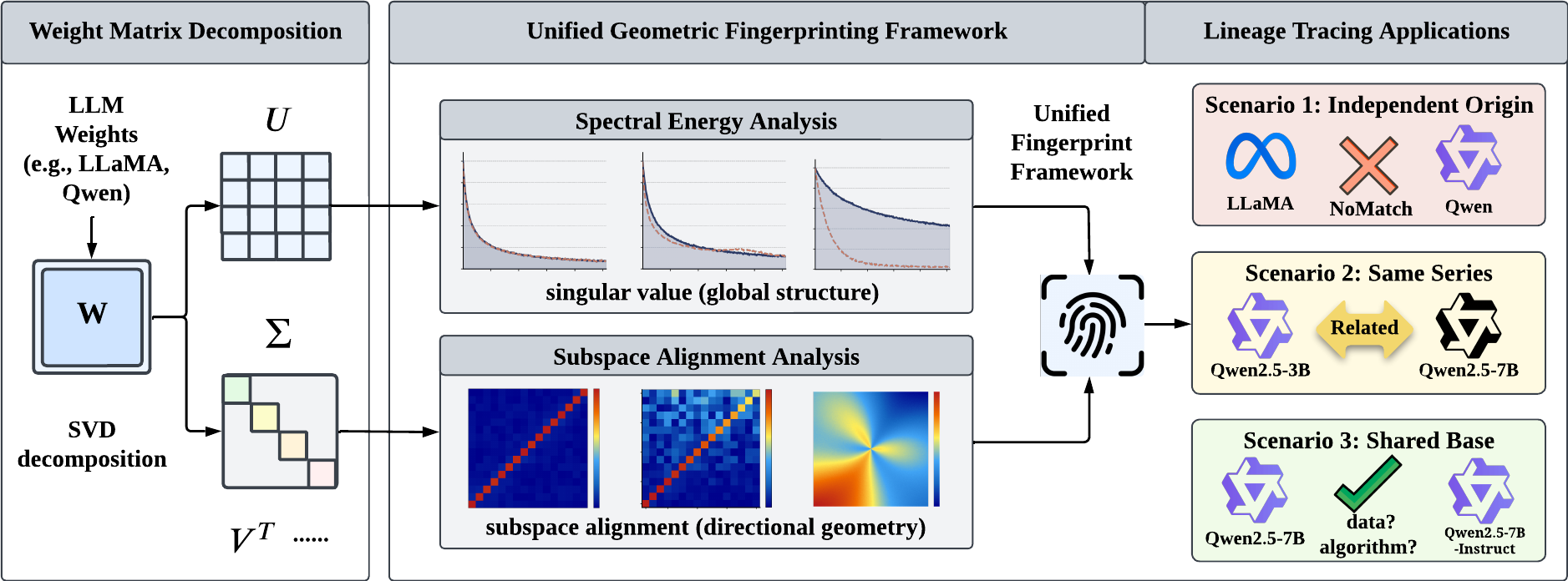}
\vspace{-2mm}
\caption{\small 
\textbf{Overview of the proposed geometric fingerprinting framework for LLM lineage analysis.}
Given model weight matrices, we apply SVD to decompose them into two complementary representations: 
(i) \textit{spectral energy} ($\Sigma$), which captures global magnitude patterns and enables coarse-grained discrimination across model families and scales, and 
(ii) \textit{subspace alignment} ($U$), which models directional relationships via subspace alignment and supports fine-grained differentiation among closely related models. 
These complementary signals are unified into a single fingerprint, revealing hierarchical model relationships and enabling robust lineage tracing in weight space.
}
\vspace{-4mm}
\label{fig:teaser}
\end{figure}

These limitations suggest that existing approaches fail to capture intrinsic structural signals that reflect model lineage.
This raises a fundamental question:

\vspace*{1mm}
\begin{tcolorbox}[before skip=2mm, after skip=0.0cm, boxsep=0.0cm, middle=0.0cm, top=0.1cm, bottom=0.1cm, boxrule=0.6pt]
\begin{center}
\textit{\textbf{(Q)} What weight-space structures serve as reliable fingerprints for LLM lineage identification?}
\end{center}
\end{tcolorbox}
\vspace*{2mm}

In this work, we study model lineage from a weight-only perspective under a white-box setting, where model parameters are accessible while training data and metadata are not~\citep{casper2024black, mokander2023auditing}. 
Such access is increasingly relevant given the growing availability of open-weight models and the need for transparent model auditing.

We formulate this problem as a \emph{hierarchical lineage discrimination} task with three regimes of increasing shared information: 
(S1) \emph{independent-origin}, where models are trained independently; 
(S2) \emph{same-series}, where same series models share architectures and training pipelines but differ primarily in parameter scale; 
and (S3) \emph{shared-base}, where models originate from a common pretrained model and diverge through post-training. 
These regimes serve as 
\begin{wraptable}{r}{0.5\textwidth}
  \vspace*{-5mm}
  \centering
\caption{\small 
Comparison of the applicability of existing methods---including Accurate Weight Matrix (AWM)~\citep{zeng2025awm}, HUman-REadable Fingerprint (HuREF)~\citep{zeng2024huref}, and Parameter Distribution Fingerprint (PDF)~\citep{yoon2025intrinsic}---and our approach across model lineage scenarios (S1-S3) and data-algorithm setups.
Here, IO, SF, and SB denote independent-origin, same-series, and shared-base scenarios. 
Data and Algo. indicate the ability to capture fine-grained variations within the shared-base regime induced by dataset scale and post-training algorithms.
}
  \vspace*{0mm}
  \label{tab:method-comparison}
  \scriptsize
  \renewcommand{\arraystretch}{1.15}
  \setlength{\tabcolsep}{4pt}
  \resizebox{\linewidth}{!}{
  \begin{tabular}{l|ccccc}
    \toprule[1pt]
    \midrule
     &  \begin{tabular}[l]{@{}l@{}} 
S1: IO
  \end{tabular}   & S2: SF &  S3: SB & Data. & Algo. \\
    \midrule
    AWM & \qmark & \qmark & \cmark & \qmark & \qmark \\
    HuREF  & \qmark & \qmark & \cmark & \qmark & \qmark \\
    PDF  & \qmark & \qmark & \cmark & \qmark & \qmark \\
    \midrule
    \rowcolor{gray!10}
    \textbf{Ours} 
          & \cmark & \cmark & \cmark & \cmark & \cmark \\
    \midrule
    \bottomrule[1pt]
  \end{tabular}
  }
  \vspace*{-2mm}
\end{wraptable}
representative reference points rather than an exhaustive taxonomy of all possible lineage relationships.
A desirable approach should therefore operate at multiple levels, separating these regimes while also characterizing intermediate relationships and distinguishing fine-grained variations within \emph{shared-base} models (\textit{e.g.}, differences induced by data scale and post-training), and generalizing across heterogeneous settings.

To address this, we propose a unified \emph{geometric fingerprinting framework} that analyzes weight matrices from two complementary perspectives, as \textbf{Fig.\,\ref{fig:teaser}} shows. 
Model relationships exhibit both coarse-grained structure across regimes and fine-grained variations within the shared-base setting; however, existing methods fail to capture both simultaneously (as summarized in \textbf{Table\,\ref{tab:method-comparison}} and will be detailed in Sec.\,\ref{sec:problem}).
We characterize \textit{spectral energy} of singular values to capture global magnitude patterns that separate model families and scales, and analyze directional geometry via \textit{subspace alignment} to reveal finer variations among closely related models. 
These perspectives are complementary: spectral energy distinguishes coarse differences across scenarios, while directional geometry remains discriminative within the shared-base setting. 
It captures variations induced by data scale, post-training methods, and their resulting geometric transformations in weight space.
In summary, our \textbf{key contributions} are as follows:

\vspace{-1mm}
\ding{172} We propose a hierarchical LLM lineage discrimination problem across three regimes, independent-origin, same-series, and shared-base, with further fine-grained discrimination within the shared-base setting, enabling coarse-to-fine analysis of model relationships.

\vspace{-1mm}
\ding{173} We show that spectral energy derived from singular values captures global magnitude patterns in weight space, enabling coarse-grained separation between independent-origin and same-series models.

\vspace{-1mm}
\ding{174} We introduce a subspace alignment analysis that captures geometric variations in weight space, enabling fine-grained discrimination among closely related shared-base models.

\vspace{-1mm}
\ding{175} 
We conduct extensive experiments on over 110 open-weight LLM pairs, showing that our approach achieves both coarse-grained regime separation and fine-grained discrimination, capturing variations induced by data scale and post-training.

\section{Related Works}
\label{sec:related-works}

\textbf{LLM Provenance and Fingerprinting.} 
The rapid growth of open-weight LLMs and increasingly diverse development pathways, including fine-tuning, alignment, distillation, model merging, among others, have made model \emph{provenance} increasingly challenging \citep{touvron2023llama, zhang2024tamm,ouyang2022training,zhang2026unleashing, bai2022constitutional,zhang2026towards,chen2026safety,chung2024scaling, wang2023self,zhang2023tile}.
While \emph{fingerprinting} may refer to embedded identifiers or behavioral signatures~\citep{nikolic2025model,kuditipudi2025blackbox}, we use it to denote passive weight-space signatures whose pairwise geometry reveals model lineage.
Following \citep{shao2025sok}, existing methods fall into black-box, white-box static, and white-box forward or backward approaches.
Black-box methods infer identity from model behavior, including natural output differences~\citep{pasquini2025llmmap,yang2024fingerprint,iourovitski2024hide,sun2025idiosyncrasies,bitton2025detecting,suzuki2025natural} and targeted query-response probing~\citep{gubri2024trap,jin2024proflingo,xu2025rap,tsai2025rofl}.
White-box static methods analyze model weights directly: HuRef~\citep{zeng2024huref} constructs algebraic attention-weight fingerprints, AWM~\citep{zeng2025awm} applies CKA with layer matching, and PDF~\citep{yoon2025intrinsic} correlates layer-wise parameter statistics.
Recent spectral methods derive transformation-invariant signatures from attention-weight compositions. GhostSpec~\citep{wang2026ghost} uses singular-value spectra, while SELF~\citep{zhang2025self} combines singular values and eigenvalues.
Both focus on binary verification between a designated source and a suspect model while preserving source identity under model transformations.
Forward methods use inference-time probing~\citep{zhang2024reef,chen2022copy,zhang2024easydetector}, whereas backward methods exploit gradients~\citep{wu2025gradient}.
In contrast, we study hierarchical lineage across independent-origin, same-series, and shared-base regimes, combining spectral energy and subspace alignment to capture coarse relationships and fine-grained post-training differences.

\textbf{Spectral and Geometric Analysis of LLM Weights.}
Spectral analysis of LLM weight matrices has emerged as an effective tool for understanding representation structure and training dynamics~\citep{martin2021predicting,martin2021implicit}. 
LLM DNA~\citep{wu2025llmdna} shows that projecting embeddings into a low-dimensional functional space suffices for lineage detection. 
Since weight matrices induce representations, identity-relevant structure should likewise be reflected in weight-space geometry.
This spectral structure is largely preserved under adaptation: LoRA parameter updates yield \textit{intruder dimensions} that exist outside the pretrained model's principal singular vectors~\citep{shuttleworth2024lora, meng2024pissa}, RLVR updates land in off-principal, low curvature subspaces~\citep{zhu2025path}. Despite these task-specific deviations, the weight spaces of models with the same architecture systematically converge to a shared, low-rank subspace~\citep{kaushik2025universal}.
\section{Tracing LLM Lineage: Motivation and Problem Statement}
\label{sec:problem}


\textbf{Weight-Space Representation of LLMs.}
In this work, we consider a white-box setting with access to pre-trained open-weight LLMs, but not their training or testing data. A transformer-based LLM is parameterized by $\theta$, represented as a collection of layer-wise weight matrices \citep{zeng2025awm, yoon2025intrinsic, zeng2024huref}:
\begin{align}
\btheta = \left\{ \mathbf W^{\scriptscriptstyle(l)}_{c} \;\middle|\; l \in [L],\; c \in \{ Q, K, V, O, FFN_{UP}, FFN_{DOWN}, FFN_{GATE} \}\right\}.
\label{eq:model_setup}
\end{align}
Here, $l \in [L] = \{1,2,\ldots, L\}$ denotes the layer index, and $c$ indexes the component type within each layer (\textit{e.g.}, $Q$, $K$, $V$, $O$ for attention projections, and corresponding FFN components). 
Given two models parametrized by $\btheta_a$ and $\btheta_b$,
we aim to characterize their (dis)similarity through these weight-space structures.

\begin{figure}[htb]
  \centering
  \vspace{-2mm}
  \begin{tabular}{@{}c@{\hspace{2mm}}c@{\hspace{2mm}}c@{}}
    \includegraphics[width=0.32\linewidth]{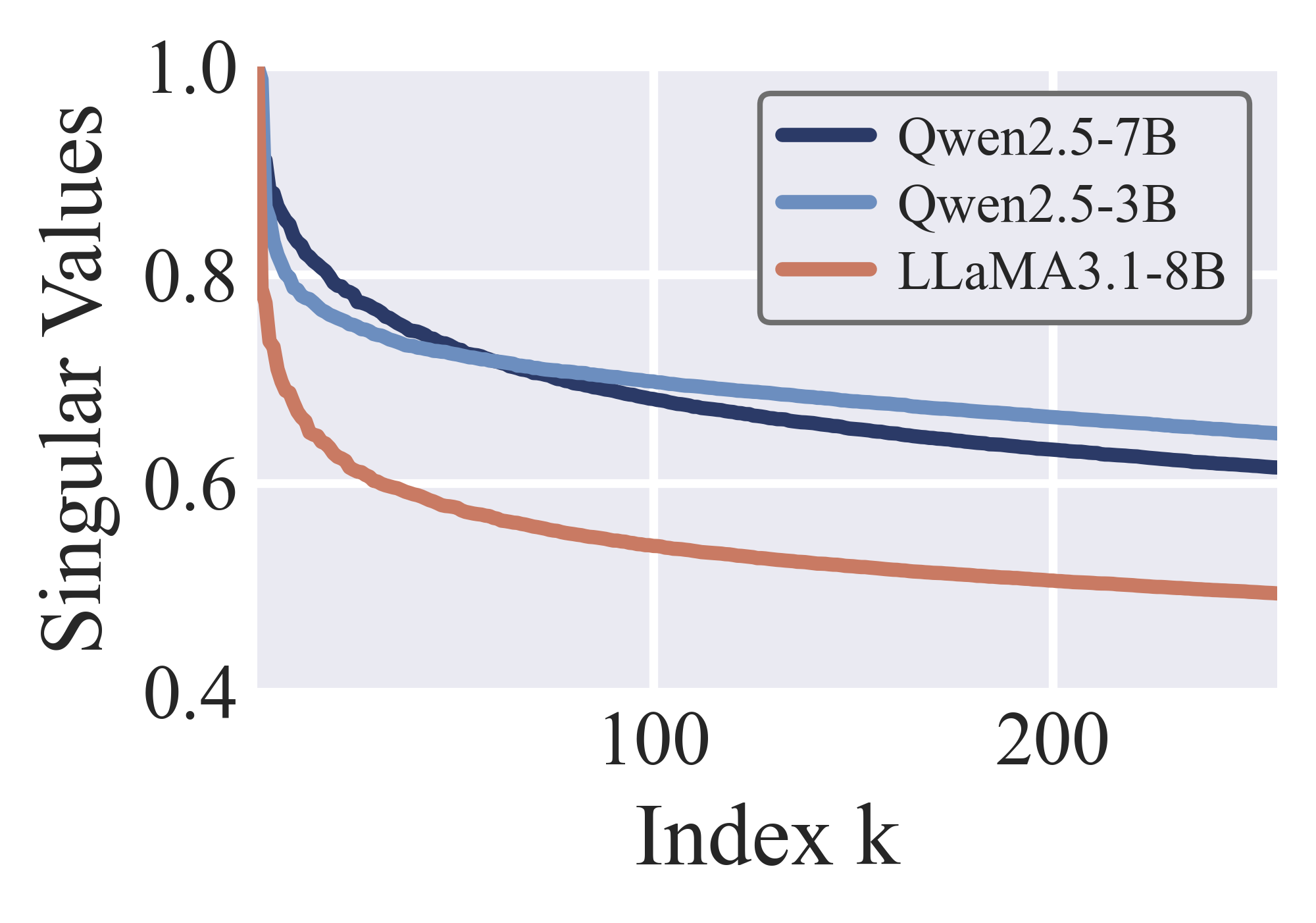} &
    \includegraphics[width=0.32\linewidth]{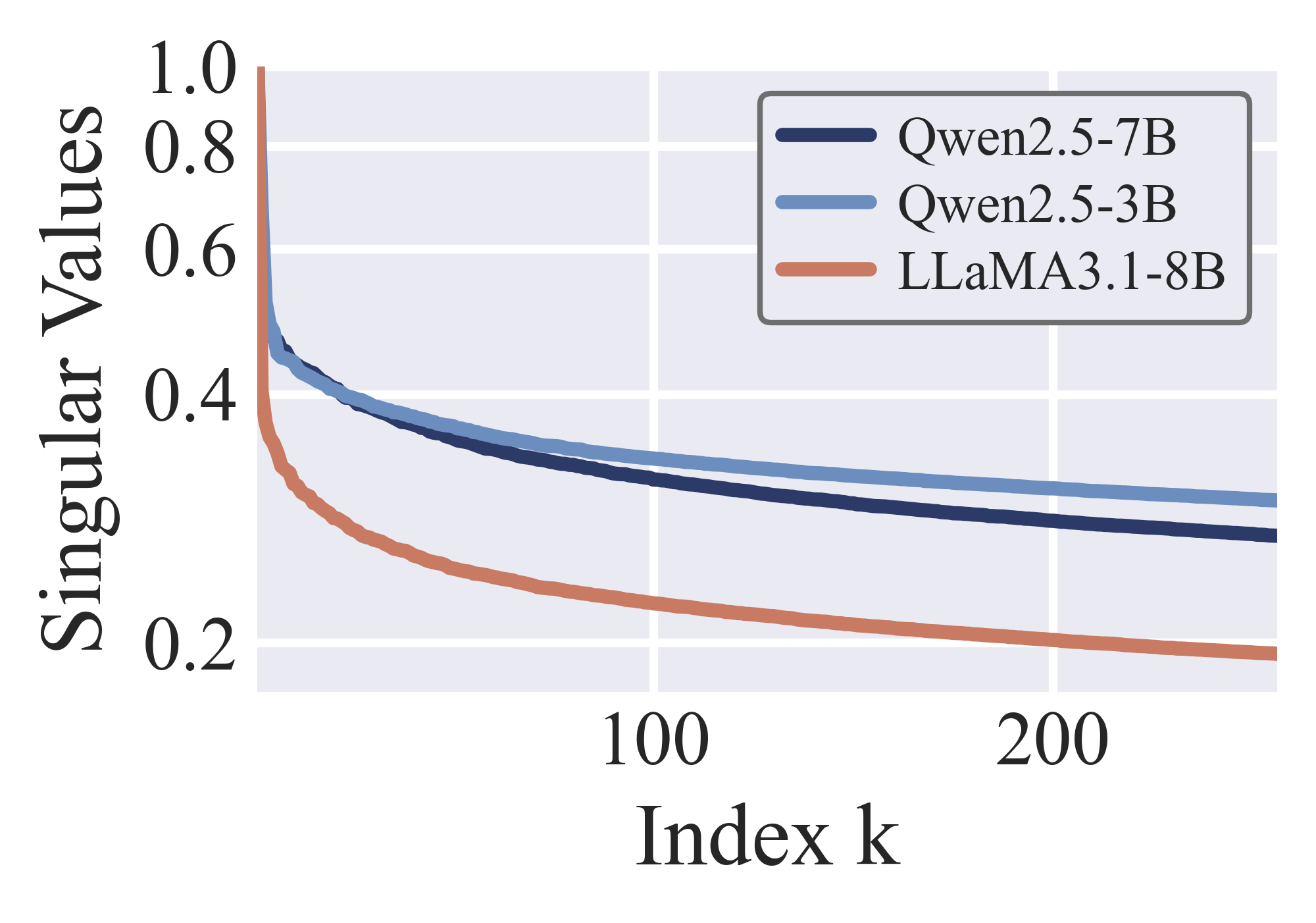} &
    \includegraphics[width=0.32\linewidth]{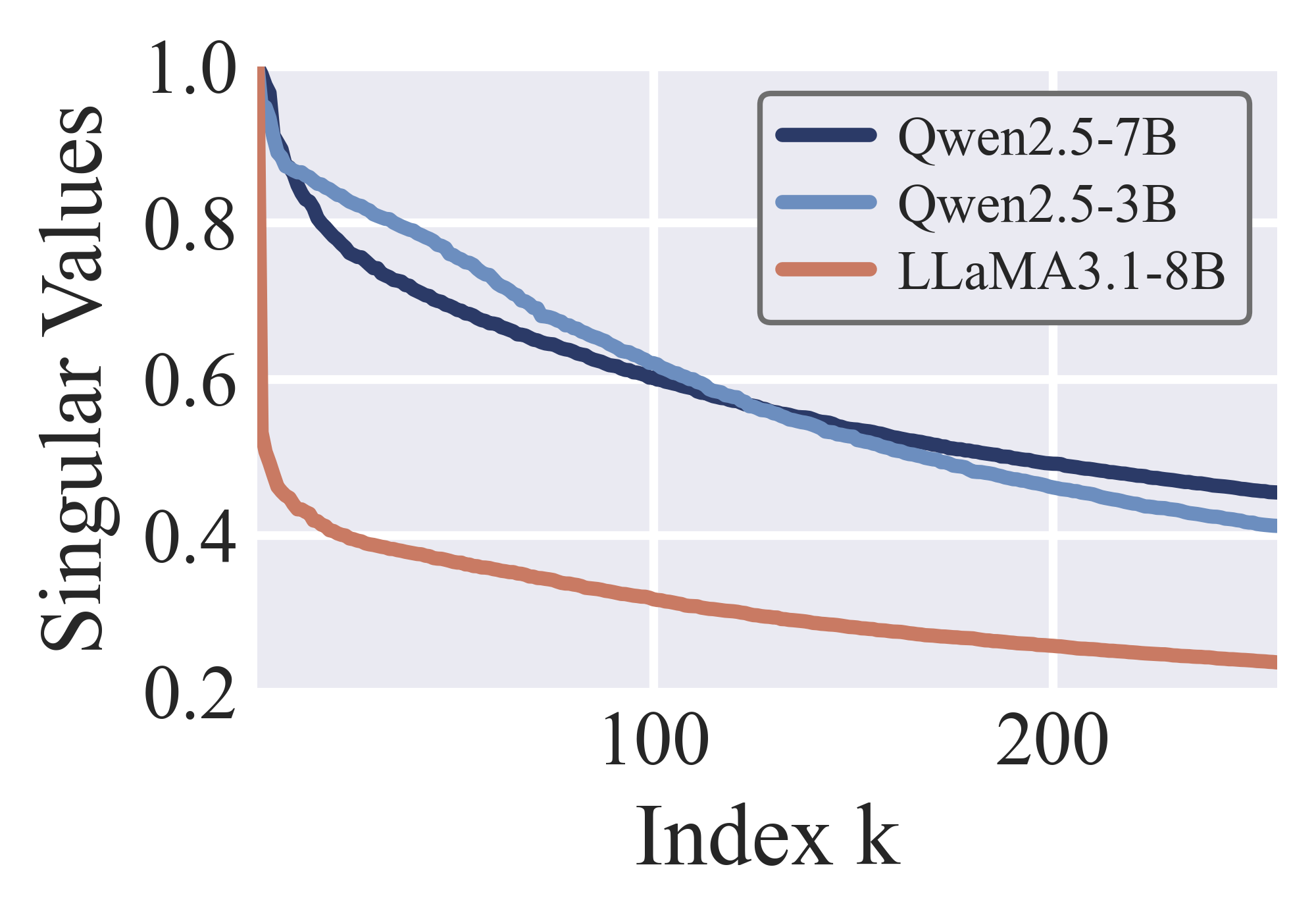} \vspace*{-2mm}\\
    \small (a) ${FFN}_{\text{UP}}$ projection &
    \small (b) ${FFN}_{\text{GATE}}$ projection &
    \small (c) $Q$ projection \\
  \end{tabular}
  \vspace{-2mm}
  \caption{\small 
Singular value distributions of middle-layer weight matrices for three component types $c$ in (a)--(c) across models (top-256 singular values, accommodating varying matrix sizes).
The x-axis denotes the singular value index $k$ (descending), and the y-axis shows normalized singular values (scaled by the largest value).
The middle layer is the $\lfloor L/2 \rfloor$-th layer of a model with $L$ layers.
Qwen2.5-3B and Qwen2.5-7B are from the same series, while LLaMA3.1-8B is from a different family.
  }
  \vspace{-2mm}
  \label{fig:svd-distribution}
\end{figure}

\textbf{Motivation: Spectral Signals in Weight Space and Limitations of Existing Methods.}
To probe structural signals in weight space, we first examine singular value spectra of weight matrices across models (\textbf{Fig.\,\ref{fig:svd-distribution}}). 
We observe consistent patterns across layers and weight types: \textit{models from the same series} (\textit{e.g.}, Qwen2.5-3B and Qwen2.5-7B) \textit{exhibit highly aligned spectral decay} in their leading singular values, whereas \textit{models from different families} (\textit{e.g.}, LLaMA3.1-8B against others) show \textit{clear deviations}. 
These differences are most pronounced in the leading singular values, which capture dominant energy and global structure. 
Additional visualization results are provided in   Appendix\,\ref{appen:spec-sig}. 
The observations in Fig.\,\ref{fig:svd-distribution} suggest that weight space encodes systematic structural signatures of model lineage, with closer models exhibiting higher spectral similarity.


\begin{wrapfigure}{r}{0.52\textwidth}
  \vspace*{-5mm}
  \centering
  \includegraphics[width=0.92\linewidth]{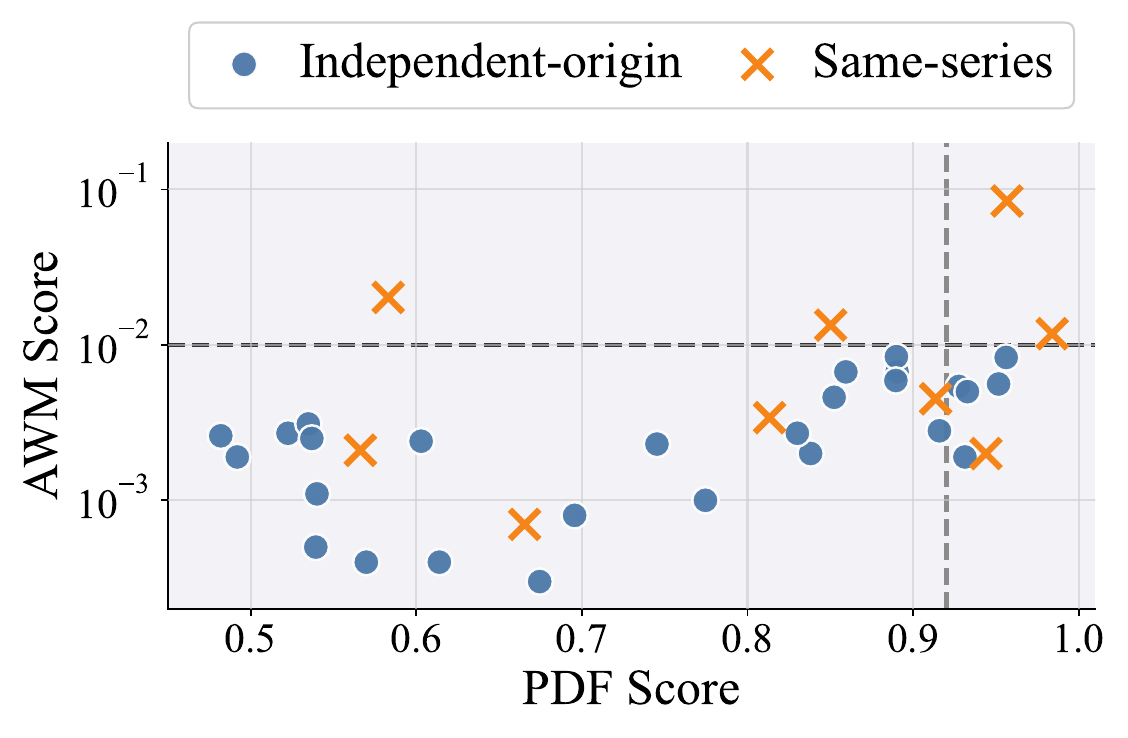}
  \vspace*{-5mm}
  \caption{\small 
Normalized similarity scores in [0, 1] for model pairs under two relationship scenarios: 
\textit{independent-origin} and \textit{same-series}.
Each point represents a model pair, positioned by AWM(y-axis) and PDF(x-axis) scores.
Dashed lines denote the decision thresholds of AWM (horizontal) and PDF (vertical), separating regimes.
  }
  \vspace*{-2mm}
  \label{fig:baseline-failure}
\end{wrapfigure}

Despite the clear spectral patterns in weight space, \textbf{existing methods fail to capture these distinctions}. 
\textbf{Fig.\,\ref{fig:baseline-failure}} shows model differentiation scores from two representative baselines: AWM (accurate weight-matrix fingerprinting)~\citep{zeng2025awm} and PDF (parameter distribution fingerprint)~\citep{yoon2025intrinsic}. We compare independent-origin pairs (\textcolor{myblue}{$\bullet$}, no shared initialization or training pipeline) with same-series pairs (\textcolor{myorange}{$\times$}, different scales sharing architecture and training pipeline) across diverse model pairs(see Appendix\,\ref{appen:model-zoo}).
Here, AWM computes centered kernel alignment (CKA) between attention weights with Hungarian-based layer matching, while PDF measures correlation of per-layer standard deviation sequences.
From the AWM perspective (y-axis), same-series pairs largely overlap with independent-origin pairs, and even with a threshold at $\text{AWM}=10^{-2}$, many are misclassified as independent-origin. 
From the PDF perspective (x-axis), separation  is even less distinct (\textit{e.g.}, at $\text{PDF}=0.92$), with substantial overlap that makes discrimination difficult.

\textbf{Proposed Hierarchical Lineage Discrimination Problem.}
As motivated by  Fig.\,\ref{fig:baseline-failure}, existing similarity metrics fail to distinguish even simple lineage relationships (\textit{e.g.}, whether models share a common origin). 
This raises a key question: \textit{Can we develop a reliable lineage discrimination metric that is both sensitive and generalizable?}

\begin{wrapfigure}{r}{0.55\textwidth}
  \vspace{-2mm}
  \centering
  \includegraphics[width=0.95\linewidth]{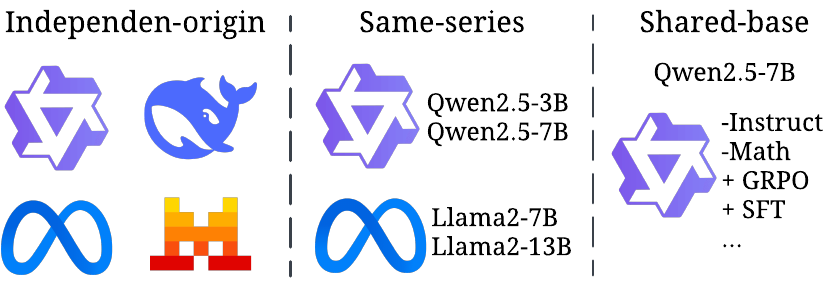}
  \vspace{-2mm}
  \caption{\small
Model relationships form three regimes with increasing degrees of shared information (left to right). 
\textit{Independent-origin:} models are trained independently without shared initialization, data, or training pipeline. 
\textit{Same-series:} models from the same series share architectures and training pipelines, but differ in scale.
\textit{Shared-base:} models are derived from a common pretrained model and diverge through different post-training data and procedures (or vs. no post-training). 
  }
  \vspace{-2mm}
  \label{fig:lineage}
\end{wrapfigure}

Model relationships arise from shared training factors, such as initialization, data, and optimization, forming a hierarchical notion of \textit{lineage}. 
As shown in \textbf{Fig.~\ref{fig:lineage}}, we define three regimes with increasing shared information: \textbf{independent-origin}, \textbf{same-series}, and \textbf{shared-base}. 
Given two models $\btheta_a$ and $\btheta_b$, we infer their relationship via a similarity function $S(\btheta_a, \btheta_b)$.
The model zoo of pairs that follow the hierarchy $S_{\text{independent-origin}} < S_{\text{same-series}} < S_{\text{shared-base}}$ is listed in Appendix\,\ref{appen:model-zoo}.

Notably, the first two regimes allow differences in architecture and scale. While such cases may appear less relevant at first glance, they serve two key purposes. 
First, a reliable similarity metric should generalize across this hierarchy: if it can distinguish shared-base models (\textit{e.g.}, post-trained variants) \citep{ouyang2022training, lambert2024tulu, guo2025deepseek}, it should also separate simpler coarse regimes with larger structural gaps. However, Fig.~\ref{fig:baseline-failure} shows that existing metrics fail to generalize even in these easier settings. 
Second, identifying model origin (\textit{e.g.}, brand or source) across heterogeneous models is itself valuable,  as models from the same source can exhibit consistent spectral signatures despite architectural differences (Fig.~\ref{fig:svd-distribution}).
In the literature, existing methods fail to separate coarse-grained regimes and only capture the distinctions between shared-base models with the others, as summarized in Table\,\ref{tab:method-comparison}.
Based on this, we adopt a coarse-to-fine perspective: global structure separates regimes, while fine-grained differences (especially within \textit{shared-base} models) are driven by data and post-training processes.

\section{Coarse-Grained Lineage Discrimination via Spectral Energy}
\label{sec:spectral}

\textbf{Spectral Trace Fingerprint.}
Given an LLM ($\btheta$) with layer-wise weight matrices defined in \eqref{eq:model_setup}, let $\mathbf W^{\scriptscriptstyle(l)}_{c} \in \mathbb{R}^{m \times n}$ have singular values $\{\sigma_i\}_{i=1}^{n}$ (assuming $n \leq m$ without loss of generality). Motivated by Fig.\,\ref{fig:svd-distribution}, we propose using  {spectral energy} as the key indicator to encode $\mathbf W^{\scriptscriptstyle(l)}_{c} $ for lineage discrimination, given by the sum of the squared  singular values, $ \sum_{i} \sigma_i^2$.

A key advantage of using spectral energy as a model fingerprint is its computational efficiency, as it can be readily obtained via the trace operation on the weights $\mathbf W^{\scriptscriptstyle(l)}_{c}$. That is, 
\begin{align}
t(\mathbf W^{\scriptscriptstyle(l)}_{c}) \Def \sqrt{\mathrm{tr}\!\left((\mathbf W^{\scriptscriptstyle(l)}_{c})^\top \mathbf W^{\scriptscriptstyle(l)}_{c}\right)} = \sqrt{\sum_{i=1}^{n}\sigma_i^2},
\label{eq:trace_W}
\end{align}
where $t(\mathbf W^{\scriptscriptstyle(l)}_{c})$ refers to   the {trace-based spectral energy}.
Based on \eqref{eq:trace_W},
we then construct layer-wise fingerprints for each component $c$ of the model $\btheta$:
\begin{align}
\tau_c(\btheta) \Def \big[ t(\mathbf W^{\scriptscriptstyle(1)}_{c}), \ldots, t(\mathbf W^{\scriptscriptstyle(L)}_{c}) \big].
\label{eq:trace_vector}
\end{align}
We refer to $\tau_c(\btheta)$ as the \textit{spectral trace fingerprint}  of component type $c$ in model $\btheta$.

\textbf{Model Lineage Discrimination Metric: Trace.}
Given the vector-wise fingerprints  in \eqref{eq:trace_vector}, we next develop a scalar-valued, easy-to-compare metric for model lineage discrimination. This metric quantifies the relationship between two models and determines their lineage category, \textit{i.e.}, independent-origin, same-series, or shared-base, as defined in Sec.\,\ref{sec:problem}.

For two models $\btheta_a$ and $\btheta_b$, 
we extract their trace fingerprints for each component type $c$, \textit{i.e.}, $\tau_c(\btheta_a)$ and $\tau_c(\btheta_b)$ from \eqref{eq:trace_vector}.
Since models may differ in depth, especially those in \textit{independent-origin} and \textit{same-series}, the resulting fingerprints can have different lengths.
To facilitate model comparison, we align the fingerprints via linear interpolation by resampling each sequence to a common length:
\begin{align}
\tilde{\tau}_c(\btheta)[j]
= \mathrm{Interp}\!\left(\tau_c(\btheta), \frac{j-1}{L_{\max}-1}\right), \quad j=1,\ldots,L_{\max},
\end{align}
where $L_{\max} = \max(L_a, L_b)$, with $L_a$ and $L_b$ denoting the numbers of layers in $\btheta_a$ and $\btheta_b$, respectively.
Here, $\mathrm{Interp}(\cdot)$ denotes piecewise linear interpolation over the normalized layer index.
For each component type, we compute the similarity between aligned trace fingerprints (\textit{i.e.}, $\tilde{\tau}_c(\btheta_a)$ and $\tilde{\tau}_c(\btheta_b)$) using Pearson correlation, yielding our proposed similarity metric, \textbf{Trace}:
\begin{align}
S_c(\btheta_a, \btheta_b)
= \mathrm{Corr}\big(\tilde{\tau}_c(\btheta_a), \tilde{\tau}_c(\btheta_b)\big)
, \quad S(\btheta_a, \btheta_b) = \mathbb E_{c} [ S_c(\btheta_a, \btheta_b) ].
\label{eq:Sc_trace}
\end{align}
Here, $S(\btheta_a, \btheta_b)$ averages similarities across all component types to produce a {scalar-valued model lineage discrimination metric}, which serves as the Trace-oriented similarity measure for determining the lineage category to which a pair of models belongs.
We refer readers to \textbf{Alg.\,\ref{alg:trace-fingerprint}} in Appendix\,\ref{appen:trace-algo} for an algorithm summary.

\begin{figure}[t]
  \vspace*{-0mm}
  \centering
  \includegraphics[width=0.9\linewidth]{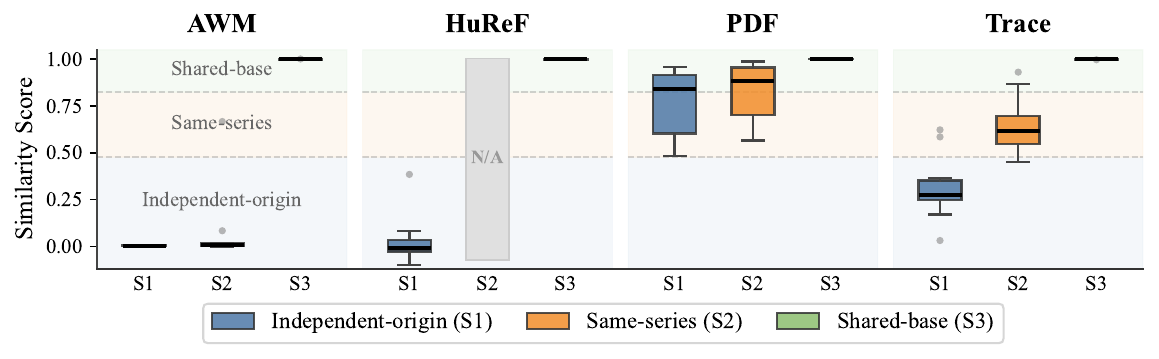}
  \vspace*{-0mm}
\caption{\small 
Distribution of Trace-based similarity scores \eqref{eq:Sc_trace} 
(normalized to [0,1]) 
for distinguishing model relationships across three scenarios: \textit{(S1) independent-origin}, \textit{(S2) same-series}, and \textit{(S3) shared-base}. 
For comparison, distributions of baseline methods (AWM, HuReF, PDF) are also shown. The x-axis denotes ground-truth lineage scenarios, each containing corresponding model pairs, while shaded regions on the y-axis indicate desired scenario-specific score ranges, with boundaries defined by the mean trace scores. Detailed settings and results are provided in Appendix\,\ref{appen:spectral-energy}.
}
  \vspace*{-3mm}
  \label{fig:box-plot-all}
\end{figure}

\textbf{Coarse-Grained Lineage Discrimination Results.}
\textbf{Fig.\,\ref{fig:box-plot-all}} illustrates the distribution of the proposed Trace scores \eqref{eq:Sc_trace} across the three model lineage discrimination scenarios: \textit{(S1) independent-origin}, \textit{(S2) same-series}, and \textit{(S3) shared-base}.
Existing methods, including AWM~\citep{zeng2025awm}, PDF~\citep{yoon2025intrinsic}, and HuReF \citep{zeng2024huref}, as well as our proposed Trace  \eqref{eq:Sc_trace}, all assign high similarity scores to shared-base models.
However, existing methods exhibit limited discriminative power in distinguishing (S1) from (S2). Specifically, AWM assigns similarly low scores to both independent-origin and same-series pairs; HuReF is not applicable to (S2) due to mismatched Transformer depths; and PDF, while improving separability, still exhibits noticeable overlap between the two scenarios.
In contrast, our method (Trace) achieves a clearer hierarchical separation across the three scenarios.
These results indicate that spectral trace captures intrinsic structural differences in weight space, leading to \textbf{Finding 1}.

\begin{center}
\vspace{-1mm}
\setlength\fboxrule{0.5pt}
\noindent\fcolorbox{black}[rgb]{0.97,0.97,0.97}{
\begin{minipage}{0.96\columnwidth}

\textbf{Finding 1 (Global Structure and Hierarchy).}
Spectral trace fingerprints \eqref{eq:trace_vector} and their associated similarity scores \eqref{eq:Sc_trace} capture global structure in weight space, enabling reliable coarse-grained separation and a clear hierarchical ordering of model lineage.
\end{minipage}}
\vspace{-1mm}
\end{center}

\begin{wrapfigure}{r}{0.52\linewidth}
\vspace{-1mm}
\centering
\includegraphics[width=0.95\linewidth]{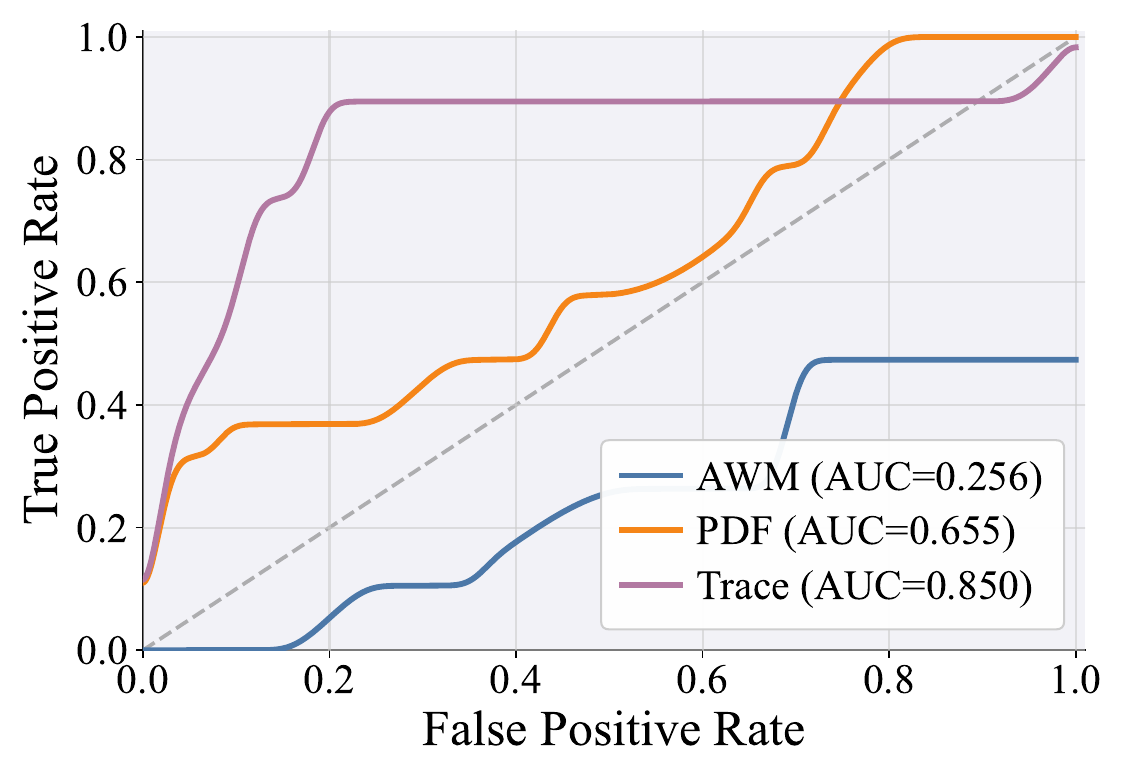}
\vspace{-4mm}
\caption{\small 
ROC curves for distinguishing \textit{independent-origin} and \textit{same-series} model pairs,
experiment setting is consistent with Fig.\,\ref{fig:box-plot-all}.
Curves are smoothed for visualization using monotonic interpolation \citep{fritsch1980monotone}, preserving AUC values.
}
\vspace{-2mm}
\label{fig:roc}
\end{wrapfigure}

Motivated by the above, we focus on the most challenging setting for existing baselines: distinguishing \textit{independent-origin} from \textit{same-series} model pairs, which exhibit minimal structural differences and are often indistinguishable by prior methods (e.g., AWM and PDF). 
\textbf{Fig.\,\ref{fig:roc}} presents ROC curves for this task. 
Existing methods show limited separability: 
AWM even performs below random chance, 
while PDF improves discrimination but still exhibits noticeable overlap. 
In contrast, our spectral trace fingerprint \eqref{eq:Sc_trace} achieves substantially higher separability, as reflected by an improved AUC. 
Additional results in Appendix\,\ref{app:adjacent-regimes} show that Trace also perfectly separates \textit{same-series} from \textit{shared-base} pairs, indicating that the improved S1 versus S2 discrimination does not compromise shared-base detection.
Together, these results demonstrate that spectral energy captures consistent structural differences across coarse-grained lineage regimes.

\begin{wrapfigure}{r}{0.52\linewidth}
\vspace{-6mm}
\centering
\includegraphics[width=0.95\linewidth]{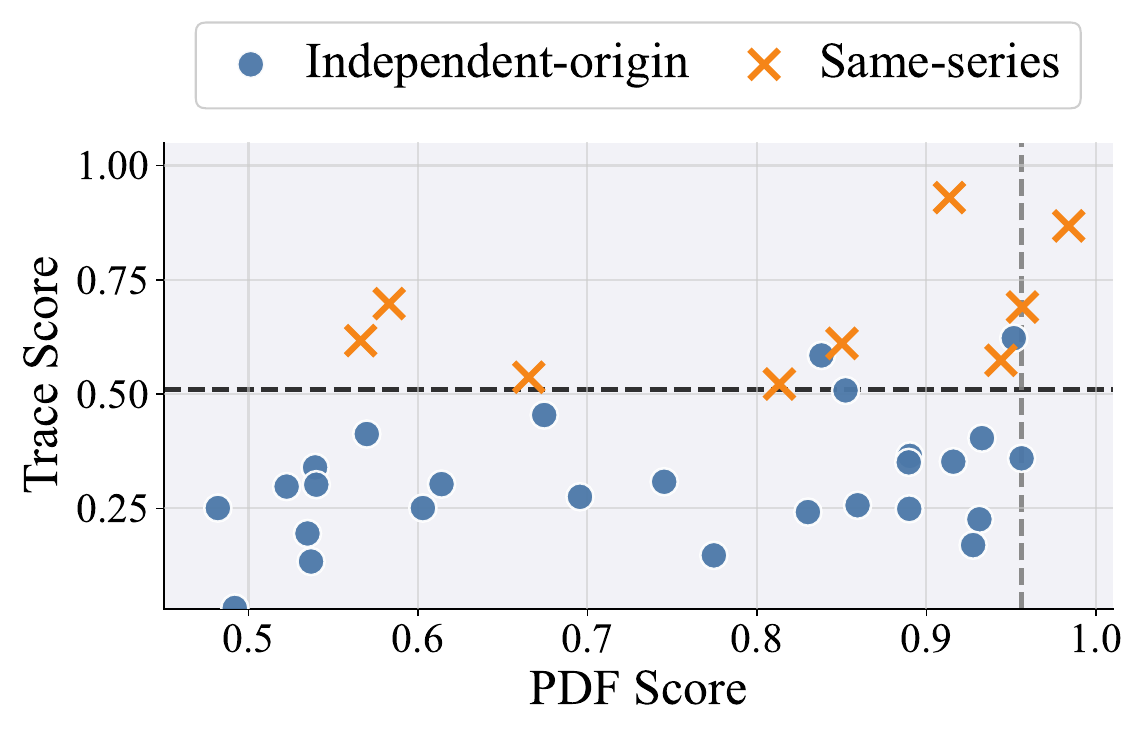}
\vspace{-2mm}
\caption{\small 
Comparison between Trace and PDF on distinguishing \textit{independent-origin} and \textit{same-series} pairs.
Experiment setup stay consistent with Fig.\,\ref{fig:baseline-failure}
}
\vspace{-5mm}
\label{fig:trace-vs-pdf}
\end{wrapfigure}

Based on the setup of Fig.\,\ref{fig:baseline-failure}, we further compare our method with the strongest baseline, PDF. 
\textbf{Fig.\,\ref{fig:trace-vs-pdf}} plots each model pair with PDF scores on the x-axis and Trace scores on the y-axis. 
While PDF yields overlapping score distributions for independent-origin and same-series pairs, our method achieves clear separation along the Trace axis. 
Through the lens of Trace, the two studied scenarios can now be separated by a consistent margin across a horizontal decision boundary, especially when compared with AWM in  Fig.\,\ref{fig:baseline-failure}, highlighting the improved discriminative power of our method.

We further examine intermediate cross-series relationships between Qwen2.5 and Qwen3 models, which share a provider and aspects of model design but differ in architecture and pretraining corpus.
As reported in \textbf{Table\,\ref{tab:cross-series}} in Appendix\,\ref{app:cross-series}, their mean Trace score of $0.479$ lies between the independent-origin and same-series average scores.
These results suggest that Trace provides a continuous relatedness signal, with the three regimes serving as representative reference points rather than exhaustive categories.

Nevertheless, Trace relies on global spectral energy and therefore provides only coarse-grained characterization.
In particular, shared-base pairs receive similarity scores close to $1$, making variations within this regime largely indistinguishable due to their highly aligned spectral patterns.
This motivates a finer-grained analysis that captures directional differences in weight space.

\section{Fine-Grained Lineage Discrimination via Subspace Alignment}
\label{sec:rotation}

\textbf{Directional Signals as Complementary Structure.}
To examine the limitation of spectral magnitude, 
we analyze a representative shared-base model pair, Qwen2.5-7B and Qwen2.5-7B-Instruct \citep{qwen2.5}, 
using the attention projection $Q$ at the 14th layer.
More visualizations are offered in Appendix\,\ref{appen:angle-signal}.
In \textbf{Fig.\,\ref{fig:rotation-motivation}(a)}, their singular value spectra nearly overlap, indicating almost identical global structure, yielding similarity scores close to 1 under Trace (and other methods) in Fig.\,\ref{fig:box-plot-all}.
However, {spectral subspace directional differences emerge}. 
In \textbf{Fig.\,\ref{fig:rotation-motivation}(b)}, while dominant subspaces remain aligned (diagonal structure), noticeable off-diagonal spread reveals local misalignment, which is measured by the correlations between groups of top-k left singular vectors from the two models across their components block. 
This is further reflected in \textbf{Fig.\,\ref{fig:rotation-motivation}(c)}, where principal angles increase, especially for higher-order components, indicating progressive subspace rotation.
These show that spectral magnitude in Sec.\,\ref{sec:spectral} captures global structure but collapses in the shared-base regime, while directional geometry provides a complementary signal that remains discriminative.

\begin{figure}[t]
\vspace{-0mm}
\centering
\setlength{\tabcolsep}{2pt}
\begin{tabular}{ccc}
    \includegraphics[width=0.32\linewidth]{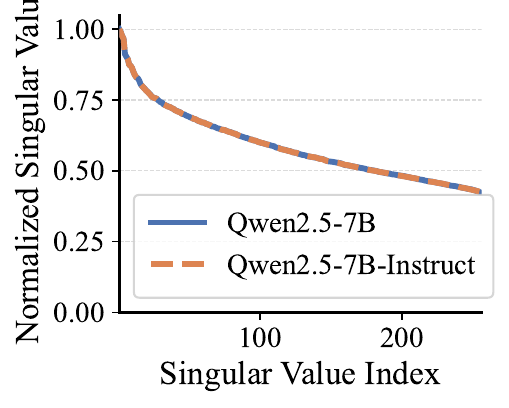} &
    \includegraphics[width=0.32\linewidth]{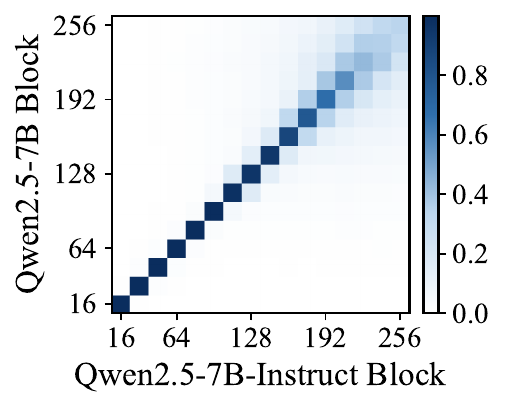} &
    \includegraphics[width=0.32\linewidth]{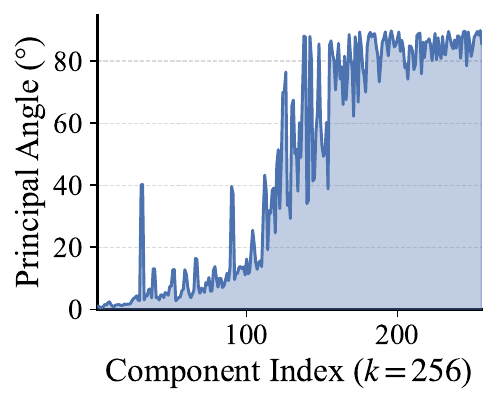} \\
    \small (a) Spectral magnitude &
    \small (b) Subspace correlation &
    \small (c) Subspace rotation \\
\end{tabular}
\vspace{-1mm}
\caption{\small
Comparison between a base model (Qwen2.5-7B) and its post-trained variant (Qwen2.5-7B-Instruct) using the attention projection $Q$ at the middle transformer layer (14th), with top-$k=256$ singular components.
(a) Spectral magnitude: normalized singular values vs. index.
(b) Subspace correlation: block-wise correlations (inner products) between groups of top-$k$ left singular vectors from the two models.
(c) Principal angles of subspace rotation: angles (in degrees) between corresponding subspaces spanned by top-$k$ singular vectors (as used in (b)), measuring subspace rotation
\citep{knyazev2002principal,zhu2012angles}.
}
\vspace{-0mm}
\label{fig:rotation-motivation}
\end{figure}

\textbf{Spectral Directional Fingerprints.}
Unlike spectral trace fingerprints, directional analysis above requires scale-aligned weight structures for subspace comparison, and is particularly suited to model lineage analysis in the \textit{shared-base} scenario. In what follows, we focus on model pairs with identical architectures but differing in training datasets and methods.

Given an LLM parameterized by $\btheta$ with layer-wise weight matrices 
$\mathbf W^{\scriptscriptstyle(l)}_{c} \in \mathbb{R}^{m \times n}$ as defined in \eqref{eq:model_setup}, 
we perform the {compact} singular value decomposition (SVD):
\begin{align}
\mathbf W^{\scriptscriptstyle(l)}_{c} = \mathbf U^{\scriptscriptstyle(l)}_{c} \Sigma^{\scriptscriptstyle(l)}_{c} ( \mathbf V^{\scriptscriptstyle(l)}_{c} )^\top,
\label{eq:rotation-finger}
\end{align}
where $\mathbf U^{\scriptscriptstyle(l)}_{c}$ and $\mathbf V^{\scriptscriptstyle(l)}_{c}$ denote the left and right singular vector subspace matrices, respectively, 
and $\boldsymbol \Sigma^{\scriptscriptstyle(l)}_{c}$ is a diagonal matrix with singular values as entries.
From \eqref{eq:rotation-finger}, we select the top-$k$ left singular vectors $\mathbf U^{\scriptscriptstyle(l)}_{c,k}$ to define a $k$-dimensional subspace that captures the dominant directions (with $k=256$ in our experiments).
For each component type $c$, we collect subspaces across layers to form a layer-wise directional fingerprint: $\big[\mathbf U^{(1)}_{c,k}, \ldots,  \mathbf U^{(L)}_{c,k} \big]$.

\textbf{Rotation-based Subspace Alignment.}
Given two models with parameters $\btheta_a$ and $\btheta_b$, 
we compare their corresponding subspaces (denoted as  $\mathbf U^{\scriptscriptstyle(l)}_{c,k}(\btheta_a)$ and $\mathbf U^{\scriptscriptstyle(l)}_{c,k}(\btheta_b)$ respectively)
at layer $l$ for component type $c$ via the \textit{cross-subspace} matrix:
\begin{align}
\mathbf C^{\scriptscriptstyle(l)}_{c}
= ( \mathbf U^{\scriptscriptstyle(l)}_{c,k}(\btheta_a) )^\top \mathbf U^{\scriptscriptstyle(l)}_{c,k}(\btheta_b).
\end{align}
Let $\{\sigma_i (\mathbf C^{\scriptscriptstyle(l)}_{c})\}_{i=1}^{k}$ denote the singular values of $\mathbf C^{\scriptscriptstyle(l)}_{c}$, 
which lie in the range $[0, 1]$ due to the orthonormality of the subspace bases~\citep{zhu2012angles,miao1992principal}. 
These singular values correspond to the cosines of the principal angles between the two subspaces, as illustrated in Fig.\,\ref{fig:rotation-motivation}(c).
We can then measure the alignment of the subspaces $\mathbf U^{\scriptscriptstyle(l)}_{c,k}(\btheta_a)$ and $\mathbf U^{\scriptscriptstyle(l)}_{c,k}(\btheta_b)$ through the singular values of $\mathbf C^{\scriptscriptstyle(l)}_{c}$. 
For example, if the subspaces are orthogonal, then $\mathbf C^{\scriptscriptstyle(l)}_{c} = \mathbf 0$. 
In contrast, large singular values indicate well-aligned subspaces, \textit{i.e.}, high similarity. 
When all singular values equal $1$, the two subspaces are identical.

Based on the above, we use the top-$J$ smallest singular values in $\{\sigma_i (\mathbf C^{\scriptscriptstyle(l)}_{c})\}_{i=1}^{k}$  to characterize the \textit{worst-case} subspace alignment, \textit{i.e.}, the largest rotation between $j$-dimensional subspaces. This leads to the layer-wise  directional similarity score:
\begin{align}
S^{\scriptscriptstyle(l)}_c(\btheta_a,\btheta_b)
= \frac{1}{J} \sum_{i \in \mathcal{I}^{\scriptscriptstyle(l)}_c} \sigma_i\!\left(\mathbf C^{\scriptscriptstyle(l)}_{c}\right) \in [0, 1],
\label{eq:Sc_subspace}
\end{align}
where $\mathcal{I}^{\scriptscriptstyle(l)}_c$ denotes the indices of the top-$J$ smallest singular values.
We use $J=3$ to aggregate a small tail of the least-aligned directions, preserving sensitivity to localized subspace rotations while reducing dependence on a single extreme direction.
A sensitivity analysis over $J$ is reported in
\textbf{Table\,\ref{tab:j-ablation}} of Appendix\,\ref{app:j-ablation}.

Since $S^{\scriptscriptstyle(l)}_c(\btheta_a,\btheta_b)$ is defined for a single component $c$ at layer $l$, we obtain a scalar discrimination score $S(\btheta_a,\btheta_b)$ (as in \eqref{eq:Sc_trace}) via hierarchical aggregation. Specifically, for each component $c$, we average the three lowest-similarity layers to obtain a component-level score, and then average across components. 
This focuses on layers exhibiting the most significant directional differences as measured by \eqref{eq:Sc_subspace}. 
The full procedure is summarized in \textbf{Alg.\,\ref{alg:rotation-fingerprint}}
, and the resulting aggregated metric based on \eqref{eq:Sc_subspace} is referred to as \textbf{subspace alignment}.

\textbf{Capturing Data-Scale Effects in Shared-base Models via Subspace Alignment.}
We consider shared-base models with varying post-training data scales, including official Instruct variants \citep{qwen2.5,yang2025qwen3,grattafiori2024llama} and controlled Alpaca-SFT models \citep{alpaca}. 
Our goal is to examine whether model lineage discrimination can be extended to a finer granularity, \textit{i.e.}, to identify differences arising from distinct data-scale (post-training) in fine-tuning.
In \textbf{Table\,\ref{tab:shared_base_sim}}, existing methods produce similarity scores close to 1.0, failing to distinguish different post-training settings. 
In contrast, the \textit{subspace alignment} metric yields lower and more diverse scores, revealing clear differences between models.

Two key insights emerge from Table\,\ref{tab:shared_base_sim}, highlighting how post-training data shape directional changes. 
\textbf{First}, official post-training variants (\textit{e.g.}, Instruct, Thinking) exhibit noticeably lower similarity than controlled Alpaca-SFT models, indicating that diverse, large-scale training data induce more substantial directional shifts in weight space. 
\textbf{Second}, within the controlled Alpaca-SFT setting, similarity decreases as the fine-tuning data scale increases from 10\% to 100\% (from 0.976 to 0.889), revealing a clear relationship between data scale and subspace alignment. 
These results demonstrate that directional similarity effectively captures data-scale effects, enabling fine-grained discrimination among shared-base models, whereas magnitude-based methods fail. 
This observation leads to \textbf{Finding 2}.

\begin{center}
\vspace{-1mm}
\setlength\fboxrule{0.5pt}
\noindent\fcolorbox{black}[rgb]{0.97,0.97,0.97}{
\begin{minipage}{0.96\columnwidth}
\textbf{Finding 2 (Sensitivity to Data Scale).}
Subspace alignment stays sensitive to post-training data scale, capturing fine-grained subspace deviations and enabling discrimination among shared-base models that differ in fine-tuning data quantity and diversity.
\end{minipage}}
\vspace{-1mm}
\end{center}

\begin{table}[htb]
\vspace*{-0mm}
\centering
\small
\caption{\small 
Similarity assessment for \textit{shared-base} model pairs under different post-training settings. 
The upper block corresponds to official variants trained with large-scale and diverse data sources, 
whereas the lower block contains controlled SFT (Supervised Fine-tuned) models trained on the dataset Alpaca using different data proportions (10\%, 25\%, 50\%, and 100\%). 
See Appendix\,\ref{appen:setup-share-base} for detailed experimental settings and more numerical results.
}
\vspace*{2mm}
\begin{tabular}{lccccc}
\toprule[1pt]
\midrule
\textbf{Model Pair} & \textbf{AWM} & \textbf{HuReF} & \textbf{PDF} & \textbf{Trace} & \textbf{Ours} \\
\midrule
Qwen2.5-7B vs Qwen2.5-7B-Instruct & 0.999 & 0.999 & 1.000 &  0.998 & 0.823\\
Qwen2.5-14B vs Qwen2.5-14B-Instruct & 0.999 & 1.000 & 1.000 &  0.998 & 0.632\\
Qwen3-4B vs Qwen3-4B-Instruct    & 0.998 & 0.999 & 0.999 & 0.998 & 0.208\\
Qwen3-4B vs Qwen3-4B-Thinking    & 0.998 & 0.999 & 0.999 & 0.999 & 0.184\\
Llama-3.1-8B vs Llama-3.1-8B-Instruct & 0.998 & 0.997 & 0.999 & 0.997 & 0.814\\
\midrule
Llama-3.1-8B vs Alpaca-SFT (10\%)     & 0.999 & 0.998 & 1.000 & 1.000 & 0.976\\
Llama-3.1-8B vs Alpaca-SFT (25\%)     & 0.999 & 0.999 & 0.999 & 0.998 & 0.935\\
Llama-3.1-8B vs Alpaca-SFT (50\%)     & 1.000 & 1.000 & 1.000 & 0.998 & 0.903\\
Llama-3.1-8B vs Alpaca-SFT (100\%)    & 1.000 & 1.000 & 1.000 & 0.998 & 0.889\\
\midrule
\bottomrule[1pt]
\vspace*{0mm}
\end{tabular}
\label{tab:shared_base_sim}
\vspace*{-0mm}
\end{table} %

\begin{figure}[htb]
\vspace{-0mm}
\centering
\includegraphics[width=0.92\linewidth]{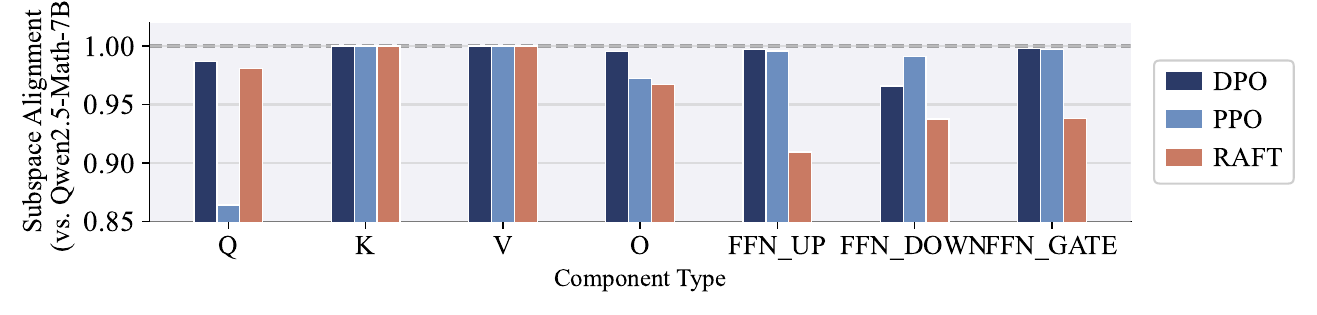}
\vspace{-0mm}
\caption{\small
Per-component subspace alignment across post-training strategies with respect to the base model Qwen2.5-Math-7B. 
Per-component subspace alignment is defined as the average of the three smallest layer-wise similarity scores \eqref{eq:Sc_subspace} across all layers, capturing worst-case directional deviation.
}
\vspace{-0mm}
\label{fig:feature-min-bar}
\end{figure}

\textbf{Algorithm-Specific Geometric Discrimination.}
Furthermore, we investigate how different post-training strategies affect weight-space geometry. 
We start from Qwen2.5-Math-7B~\citep{yang2024qwen2} and apply different post-training approaches, including PPO~\citep{schulman2017proximal}, DPO~\citep{rafailov2023direct}, and reward-ranked fine-tuning (RAFT)~\citep{dong2023raft} within the Online-DPO-R1 framework~\citep{zhangonline}. 
All methods share the same initialization and training dataset, with aligned optimization processes.
\textbf{Fig.\,\ref{fig:feature-min-bar}} shows the sensitivity of structural similarity (measured by subspace alignment scores) to different post-training strategies across model components. 
The $K$ and $V$ projections remain close to 1.0, indicating largely preserved subspaces. 
In contrast, other components exhibit method-dependent deviations, reflecting how post-training perturbs weight-space alignment. 
RAFT yields lower similarity in $FFN$ components, suggesting stronger perturbations, while PPO causes a noticeable drop in the $Q$ projection. 
By contrast, DPO maintains consistently high similarity across components. 
Notably, the $Q$ projection shows the largest variation across methods, highlighting its sensitivity to post-training dynamics and its role in capturing fine-grained geometric changes. 
This leads to \textbf{Finding 3}.

\begin{center}
\vspace{-0mm}
\setlength\fboxrule{0.5pt}
\noindent\fcolorbox{black}[rgb]{0.97,0.97,0.97}{
\begin{minipage}{0.96\columnwidth}
\textbf{Finding 3 (Algorithm-Specific Subspace Sensitivity).}
Post-training can induce algorithm-dependent directional deviations, as measured by subspace alignment, revealing distinct geometric signatures in different model components.
\end{minipage}}
\vspace{-0mm}
\end{center}

\textbf{Additional Experiments.}
Detailed numerical results on fine-grained discrimination on both data-scale and algorithmic variants are presented in Appendix\,\ref{appen:exp-fine-grain}.
We further evaluate robustness to quantization, unstructured magnitude pruning, model merging, and data distillation in Appendix~\ref{app:transformation-robustness}.
Across these transformations, Trace preserves the coarse-grained shared-base lineage signal, while Subspace Alignment captures transformation severity and component-specific directional changes.

\vspace{-0mm}
\section{Conclusion}
\label{sec:conclusion}
\vspace{0mm}
In this work, we study LLM lineage from a weight-based perspective and examine whether model identity and relationships can be inferred from intrinsic signatures. 
We formulate this as a hierarchical discrimination task and propose a unified geometric fingerprinting framework combining spectral and directional analysis to capture coarse-grained relationships across regimes and fine-grained variations within shared-base models. 
Our results show that spectral structure enables coarse-grained separation across model families, while directional geometry captures fine-grained variations induced by data scale and post-training.

\section*{Acknowledgment}
This work was supported in part by the U.S. National Science Foundation (NSF) under CISE Core Award IIS-2504263 and NSF CAREER Award IIS-2338068, the Coefficient Giving AI Safety Research Award, and the Schmidt Sciences Trustworthy AI Award.


\newpage
\section*{Ethics Statement}
\label{sec:statement}
This work proposes a passive, white-box method for analyzing LLM lineage through weight-space geometry. The method operates solely on publicly released model weights and does not involve human subjects or personal data. All models examined are open-weight and publicly available. Spectral fingerprinting is designed to support model provenance verification, intellectual-property governance, and supply-chain transparency in the rapidly expanding open-weight LLM ecosystem. As derivative models proliferate through fine-tuning, distillation, and alignment, reliable lineage attribution can help maintainers, regulators, and downstream users understand the origins of a given model and assess associated risks.

\section*{Reproducibility Statement}
We take several steps to facilitate reproducibility. All algorithmic procedures, spectral trace fingerprinting in Sec.\,\ref{sec:spectral} and subspace alignment via principal angles in Sec.\,\ref{sec:rotation}, are described with full mathematical detail, including similarity metrics, aggregation schemes, and hyperparameter choices both in the main paper and Appendix. 
Every model used in our experiments is publicly available on the HuggingFace Hub, and no proprietary data or training runs are required; fingerprint extraction operates directly on released weight files. 
The core computations rely on standard linear algebra routines. 
We report results across multiple model families, scales, and post-training algorithms to demonstrate robustness. 

\bibliography{colm2026_conference}

@article{zeng2025awm,
  title={AWM: Accurate Weight-Matrix Fingerprint for Large Language Models},
  author={Zeng, Boyi and Chen, Lin and He, Ziwei and Wang, Xinbing and Lin, Zhouhan},
  journal={arXiv preprint arXiv:2510.06738},
  year={2025}
}

@article{yoon2025intrinsic,
  title={Intrinsic Fingerprint of LLMs: Continue Training is NOT All You Need to Steal A Model!},
  author={Yoon, Do-hyeon and Chun, Minsoo and Allen, Thomas and M{\"u}ller, Hans and Wang, Min and Sharma, Rajesh},
  journal={arXiv preprint arXiv:2507.03014},
  year={2025}
}

@article{zeng2024huref,
  title={Huref: Human-readable fingerprint for large language models},
  author={Zeng, Boyi and Wang, Lizheng and Hu, Yuncong and Xu, Yi and Zhou, Chenghu and Wang, Xinbing and Yu, Yu and Lin, Zhouhan},
  journal={NeurIPS},
  year={2024}
}

@article{shao2025sok,
  title={Sok: Large language model copyright auditing via fingerprinting},
  author={Shao, Shuo and Li, Yiming and He, Yu and Yao, Hongwei and Yang, Wenyuan and Tao, Dacheng and Qin, Zhan},
  journal={arXiv preprint arXiv:2508.19843},
  year={2025}
}

@article{zhang2024reef,
  title={Reef: Representation encoding fingerprints for large language models},
  author={Zhang, Jie and Liu, Dongrui and Qian, Chen and Zhang, Linfeng and Liu, Yong and Qiao, Yu and Shao, Jing},
  journal={arXiv preprint arXiv:2410.14273},
  year={2024}
}

@article{wu2025llmdna,
  title={LLM DNA: Tracing Model Evolution via Functional Representations},
  author={Wu, Zhaomin and Zhao, Haodong and Wang, Ziyang and Guo, Jizhou and Wang, Qian and He, Bingsheng},
  journal={arXiv preprint arXiv:2509.24496},
  year={2025}
}

@misc{qwen2.5,
    title = {Qwen2.5: A Party of Foundation Models},
    url = {https://qwenlm.github.io/blog/qwen2.5/},
    author = {Qwen},
    month = {September},
    year = {2024}
}

@article{ouyang2022training,
  title={Training language models to follow instructions with human feedback},
  author={Ouyang, Long and Wu, Jeffrey and Jiang, Xu and Almeida, Diogo and Wainwright, Carroll and Mishkin, Pamela and Zhang, Chong and Agarwal, Sandhini and Slama, Katarina and Ray, Alex and others},
  journal={NeurIPS},
  year={2022}
}

@article{lambert2024tulu,
  title={Tulu 3: Pushing frontiers in open language model post-training},
  author={Lambert, Nathan and Morrison, Jacob and Pyatkin, Valentina and Huang, Shengyi and Ivison, Hamish and Brahman, Faeze and Miranda, Lester James V and Liu, Alisa and Dziri, Nouha and Lyu, Shane and others},
  journal={arXiv preprint arXiv:2411.15124},
  year={2024}
}

@article{guo2025deepseek,
  title={Deepseek-r1: Incentivizing reasoning capability in llms via reinforcement learning},
  author={Guo, Daya and Yang, Dejian and Zhang, Haowei and Song, Junxiao and Wang, Peiyi and Zhu, Qihao and Xu, Runxin and Zhang, Ruoyu and Ma, Shirong and Bi, Xiao and others},
  journal={arXiv preprint arXiv:2501.12948},
  year={2025}
}

@inproceedings{wang2026ghost,
  title={Ghost in the Transformer: Detecting Model Reuse with Invariant Spectral Signatures},
  author={Wang, Suqing and Ma, Ziyang and Xinyi, Li and Li, Zuchao},
  booktitle={AAAI},
  year={2026}
}

@article{zhang2025self,
  title={SELF: A Robust Singular Value and Eigenvalue Approach for LLM Fingerprinting},
  author={Zhang, Hanxiu and Zheng, Yue},
  journal={arXiv preprint arXiv:2512.03620},
  year={2025}
}

@article{fritsch1980monotone,
  title={Monotone piecewise cubic interpolation},
  author={Fritsch, Frederick N and Carlson, Ralph E},
  journal={SIAM Journal on Numerical Analysis},
  volume={17},
  number={2},
  pages={238--246},
  year={1980},
  publisher={SIAM}
}

@article{yang2024qwen2,
  title={Qwen2. 5-math technical report: Toward mathematical expert model via self-improvement},
  author={Yang, An and Zhang, Beichen and Hui, Binyuan and Gao, Bofei and Yu, Bowen and Li, Chengpeng and Liu, Dayiheng and Tu, Jianhong and Zhou, Jingren and Lin, Junyang and others},
  journal={arXiv preprint arXiv:2409.12122},
  year={2024}
}

@article{zhangonline,
  title={Online-dpo-r1: Unlocking effective reasoning without the ppo overhead},
  author={Zhang, Hanning and Yao, Jiarui and Ye, Chenlu and Xiong, Wei and Zhang, Tong},
  journal={Notion Blog},
  year={2025}
}

@article{grattafiori2024llama,
  title={The llama 3 herd of models},
  author={Grattafiori, Aaron and Dubey, Abhimanyu and Jauhri, Abhinav and Pandey, Abhinav and Kadian, Abhishek and Al-Dahle, Ahmad and Letman, Aiesha and Mathur, Akhil and Schelten, Alan and Vaughan, Alex and others},
  journal={arXiv preprint arXiv:2407.21783},
  year={2024}
}

@misc{alpaca,
  author = {Rohan Taori and Ishaan Gulrajani and Tianyi Zhang and Yann Dubois and Xuechen Li and Carlos Guestrin and Percy Liang and Tatsunori B. Hashimoto },
  title = {Stanford Alpaca: An Instruction-following LLaMA model},
  year = {2023},
  publisher = {GitHub},
  journal = {GitHub repository},
  howpublished = {\url{https://github.com/tatsu-lab/stanford_alpaca}},
}

@article{yang2025qwen3,
  title={Qwen3 technical report},
  author={Yang, An and Li, Anfeng and Yang, Baosong and Zhang, Beichen and Hui, Binyuan and Zheng, Bo and Yu, Bowen and Gao, Chang and Huang, Chengen and Lv, Chenxu and others},
  journal={arXiv preprint arXiv:2505.09388},
  year={2025}
}

@article{schulman2017proximal,
  title={Proximal policy optimization algorithms},
  author={Schulman, John and Wolski, Filip and Dhariwal, Prafulla and Radford, Alec and Klimov, Oleg},
  journal={arXiv preprint arXiv:1707.06347},
  year={2017}
}

@article{rafailov2023direct,
  title={Direct preference optimization: Your language model is secretly a reward model},
  author={Rafailov, Rafael and Sharma, Archit and Mitchell, Eric and Manning, Christopher D and Ermon, Stefano and Finn, Chelsea},
  journal={NeurIPS},
  year={2023}
}

@inproceedings{chen2026unlearning,
  title={Unlearning Isn't Invisible: Detecting Unlearning Traces in LLMs from Model Outputs},
  author={Chen, Yiwei and Pal, Soumyadeep and Zhang, Yimeng and Qu, Qing and Liu, Sijia},
  booktitle={International Conference on Learning Representations},
  volume={2026},
  pages={111444--111470},
  year={2026}
}

@inproceedings{chen2026safety,
  title={Safety mirage: How spurious correlations undermine vlm safety fine-tuning and can be mitigated by machine unlearning},
  author={Chen, Yiwei and Yao, Yuguang and Zhang, Yihua and Shen, Bingquan and Liu, Gaowen and Liu, Sijia},
  booktitle={International Conference on Learning Representations},
  volume={2026},
  pages={78527--78547},
  year={2026}
}

@inproceedings{zhang2023tile,
  title={Tile classification based viewport prediction with multi-modal fusion transformer},
  author={Zhang, Zhiahao and Chen, Yiwei and Zhang, Weizhan and Yan, Caixia and Zheng, Qinghua and Wang, Qi and Chen, Wangdu},
  booktitle={ACM-MM},
  year={2023}
}

@inproceedings{zhang2024tamm,
  title={Tamm: Triadapter multi-modal learning for 3d shape understanding},
  author={Zhang, Zhihao and Cao, Shengcao and Wang, Yu-Xiong},
  booktitle={CVPR},
  year={2024}
}

@article{zhang2026towards,
  title={Towards intrinsic-aware monocular 3d object detection},
  author={Zhang, Zhihao and Kumar, Abhinav and Liu, Xiaoming},
  journal={arXiv preprint arXiv:2603.27059},
  year={2026}
}

@inproceedings{zhang2026unleashing,
  title={Unleashing the power of chain-of-prediction for monocular 3d object detection},
  author={Zhang, Zhihao and Kumar, Abhinav and Ganesan, Girish Chandar and Liu, Xiaoming},
  booktitle={CVPR},
  year={2026}
}

@article{shang2025forgetting,
  title={Forgetting to forget: Attention sink as a gateway for backdooring llm unlearning},
  author={Shang, Bingqi and Chen, Yiwei and Zhang, Yihua and Shen, Bingquan and Liu, Sijia},
  journal={arXiv preprint arXiv:2510.17021},
  year={2025}
}

@article{bai2022constitutional,
  title={Constitutional ai: Harmlessness from ai feedback},
  author={Bai, Yuntao and Kadavath, Saurav and Kundu, Sandipan and Askell, Amanda and Kernion, Jackson and Jones, Andy and Chen, Anna and Goldie, Anna and Mirhoseini, Azalia and McKinnon, Cameron and others},
  journal={arXiv preprint arXiv:2212.08073},
  year={2022}
}

@article{touvron2023llama,
  title={Llama: Open and efficient foundation language models. arXiv 2023},
  author={Touvron, Hugo and Lavril, Thibaut and Izacard, Gautier and Martinet, Xavier and Lachaux, Marie-Anne and Lacroix, Timoth{\'e}e and Rozi{\`e}re, Baptiste and Goyal, Naman and Hambro, Eric and Azhar, Faisal and others},
  journal={arXiv preprint arXiv:2302.13971},
  volume={10},
  year={2023}
}

@article{team2024gemma,
  title={Gemma: Open models based on gemini research and technology},
  author={Gemma, Gemma and Mesnard, Thomas and Hardin, Cassidy and Dadashi, Robert and Bhupatiraju, Surya and Pathak, Shreya and Sifre, Laurent and Rivi{\`e}re, Morgane and Kale, Mihir Sanjay and Love, Juliette and others},
  journal={arXiv preprint arXiv:2403.08295},
  year={2024}
}

@article{chung2024scaling,
  title={Scaling instruction-finetuned language models},
  author={Chung, Hyung Won and Hou, Le and Longpre, Shayne and Zoph, Barret and Tay, Yi and Fedus, William and Li, Yunxuan and Wang, Xuezhi and Dehghani, Mostafa and Brahma, Siddhartha and others},
  journal={JMLR},
  year={2024}
}

@inproceedings{wang2023self,
  title={Self-instruct: Aligning language models with self-generated instructions},
  author={Wang, Yizhong and Kordi, Yeganeh and Mishra, Swaroop and Liu, Alisa and Smith, Noah A and Khashabi, Daniel and Hajishirzi, Hannaneh},
  booktitle={ACL},
  year={2023}
}

@article{zhou2023lima,
  title={Lima: Less is more for alignment},
  author={Zhou, Chunting and Liu, Pengfei and Xu, Puxin and Iyer, Srinivasan and Sun, Jiao and Mao, Yuning and Ma, Xuezhe and Efrat, Avia and Yu, Ping and Yu, Lili and others},
  journal={NeurIPS},
  year={2023}
}

@inproceedings{carlini2021extracting,
  title={Extracting training data from large language models},
  author={Carlini, Nicholas and Tramer, Florian and Wallace, Eric and Jagielski, Matthew and Herbert-Voss, Ariel and Lee, Katherine and Roberts, Adam and Brown, Tom and Song, Dawn and Erlingsson, Ulfar and others},
  booktitle={USENIX Security 21},
  year={2021}
}

@article{pahune2025importance,
  title={The importance of AI data governance in large language models},
  author={Pahune, Saurabh and Akhtar, Zahid and Mandapati, Venkatesh and Siddique, Kamran},
  journal={Big Data and Cognitive Computing},
  volume={9},
  number={6},
  pages={147},
  year={2025},
}

@article{kuditipudi2025blackbox,
  title={Blackbox model provenance via palimpsestic membership inference},
  author={Kuditipudi, Rohith and Huang, Jing and Zhu, Sally and Yang, Diyi and Potts, Christopher and Liang, Percy},
  journal={arXiv preprint arXiv:2510.19796},
  year={2025}
}

@article{nikolic2025model,
  title={Model provenance testing for large language models},
  author={Nikolic, Ivica and Baluta, Teodora and Saxena, Prateek},
  journal={arXiv preprint arXiv:2502.00706},
  year={2025}
}

@article{tsai2025rofl,
  title={Rofl: Robust fingerprinting of language models},
  author={Tsai, Yun-Yun and Guo, Chuan and Yang, Junfeng and van der Maaten, Laurens},
  journal={arXiv preprint arXiv:2505.12682},
  year={2025}
}

@inproceedings{casper2024black,
  title={Black-box access is insufficient for rigorous ai audits},
  author={Casper, Stephen and Ezell, Carson and Siegmann, Charlotte and Kolt, Noam and Curtis, Taylor Lynn and Bucknall, Benjamin and Haupt, Andreas and Wei, Kevin and Scheurer, J{\'e}r{\'e}my and Hobbhahn, Marius and others},
  booktitle={Proceedings of the 2024 ACM Conference on Fairness, Accountability, and Transparency},
  year={2024}
}

@article{mokander2023auditing,
  title={Auditing of AI: Legal, ethical and technical approaches},
  author={M{\"o}kander, Jakob},
  journal={Digital Society},
  volume={2},
  number={3},
  pages={49},
  year={2023},
  publisher={Springer}
}

@inproceedings{chen2022copy,
  title={Copy, right? a testing framework for copyright protection of deep learning models},
  author={Chen, Jialuo and Wang, Jingyi and Peng, Tinglan and Sun, Youcheng and Cheng, Peng and Ji, Shouling and Ma, Xingjun and Li, Bo and Song, Dawn},
  booktitle={IEEE symposium on security and privacy (SP)},
  pages={824--841},
  year={2022}
}

@article{wu2025gradient,
  title={Gradient-based model fingerprinting for llm similarity detection and family classification},
  author={Wu, Zehao and Zhao, Yanjie and Wang, Haoyu},
  journal={arXiv preprint arXiv:2506.01631},
  year={2025}
}

@inproceedings{pasquini2025llmmap,
  title={LLMmap: Fingerprinting for large language models},
  author={Pasquini, Dario and Kornaropoulos, Evgenios M and Ateniese, Giuseppe},
  booktitle={USENIX Security 25},
  pages={299--318},
  year={2025}
}

@article{yang2024fingerprint,
  title={A fingerprint for large language models},
  author={Yang, Zhiguang and Wu, Hanzhou},
  journal={arXiv preprint arXiv:2407.01235},
  year={2024}
}

@article{iourovitski2024hide,
  title={Hide and seek: Fingerprinting large language models with evolutionary learning},
  author={Iourovitski, Dmitri and Sharma, Sanat and Talwar, Rakshak},
  journal={arXiv preprint arXiv:2408.02871},
  year={2024}
}

@article{sun2025idiosyncrasies,
  title={Idiosyncrasies in large language models},
  author={Sun, Mingjie and Yin, Yida and Xu, Zhiqiu and Kolter, J Zico and Liu, Zhuang},
  journal={arXiv preprint arXiv:2502.12150},
  year={2025}
}

@article{bitton2025detecting,
  title={Detecting stylistic fingerprints of large language models},
  author={Bitton, Yehonatan and Bitton, Elad and Nisan, Shai},
  journal={arXiv preprint arXiv:2503.01659},
  year={2025}
}

@article{suzuki2025natural,
  title={Natural fingerprints of large language models},
  author={Suzuki, Teppei and Ri, Ryokan and Takase, Sho},
  journal={arXiv preprint arXiv:2504.14871},
  year={2025}
}

@inproceedings{gubri2024trap,
  title={Trap: Targeted random adversarial prompt honeypot for black-box identification},
  author={Gubri, Martin and Ulmer, Dennis and Lee, Hwaran and Yun, Sangdoo and Oh, Seong Joon},
  booktitle={Findings of ACL},
  year={2024}
}

@article{zhu2012angles,
  title={Angles between subspaces and their tangents},
  author={Zhu, Peizhen and Knyazev, Andrew V},
  journal={arXiv preprint arXiv:1209.0523},
  year={2012}
}

@article{knyazev2002principal,
  title={Principal angles between subspaces in an A-based scalar product: algorithms and perturbation estimates},
  author={Knyazev, Andrew V and Argentati, Merico E},
  journal={SIAM Journal on Scientific Computing},
  volume={23},
  number={6},
  pages={2008--2040},
  year={2002}
}

@inproceedings{jin2024proflingo,
  title={Proflingo: A fingerprinting-based intellectual property protection scheme for large language models},
  author={Jin, Heng and Zhang, Chaoyu and Shi, Shanghao and Lou, Wenjing and Hou, Y Thomas},
  booktitle={IEEE Conference on Communications and Network Security (CNS)},
  pages={1--9},
  year={2024}
}

@article{xu2025rap,
  title={Rap-sm: Robust adversarial prompt via shadow models for copyright verification of large language models},
  author={Xu, Zhenhua and Wang, Zhebo and Li, Maike and Xing, Wenpeng and Hu, Chunqiang and Zhi, Chen and Han, Meng},
  journal={arXiv preprint arXiv:2505.06304},
  year={2025}
}

@article{miao1992principal,
  title={On principal angles between subspaces in Rn},
  author={Miao, Jianming and Ben-Israel, Adi},
  journal={Linear algebra and its applications},
  volume={171},
  pages={81--98},
  year={1992},
  publisher={Elsevier}
}

@INPROCEEDINGS{zhang2024easydetector,
  author={Zhang, Jie and Li, Jiayuan and Fei, Haiqiang and Li, Lun and Zhu, Hongsong},
  booktitle={2024 IEEE 23rd International Conference on Trust, Security and Privacy in Computing and Communications (TrustCom)}, 
  title={EasyDetector: Using Linear Probe to Detect the Provenance of Large Language Models}, 
  year={2024},
  pages={2410-2417}}

@article{shuttleworth2024lora,
  title={Lora vs full fine-tuning: An illusion of equivalence},
  author={Shuttleworth, Reece and Andreas, Jacob and Torralba, Antonio and Sharma, Pratyusha},
  journal={arXiv preprint arXiv:2410.21228},
  year={2024}
}

@article{meng2024pissa,
  title={Pissa: Principal singular values and singular vectors adaptation of large language models},
  author={Meng, Fanxu and Wang, Zhaohui and Zhang, Muhan},
  journal={NeurIPS},
  year={2024}
}

@article{zhu2025path,
  title={The path not taken: Rlvr provably learns off the principals},
  author={Zhu, Hanqing and Zhang, Zhenyu and Huang, Hanxian and Su, DiJia and Liu, Zechun and Zhao, Jiawei and Fedorov, Igor and Pirsiavash, Hamed and Sha, Zhizhou and Lee, Jinwon and others},
  journal={arXiv preprint arXiv:2511.08567},
  year={2025}
}

@article{kaushik2025universal,
  title={The universal weight subspace hypothesis},
  author={Kaushik, Prakhar and Chaudhari, Shravan and Vaidya, Ankit and Chellappa, Rama and Yuille, Alan},
  journal={arXiv preprint arXiv:2512.05117},
  year={2025}
}

@article{martin2021predicting,
  title={Predicting trends in the quality of state-of-the-art neural networks without access to training or testing data},
  author={Martin, Charles H and Peng, Tongsu and Mahoney, Michael W},
  journal={Nature Communications},
  volume={12},
  number={1},
  pages={4122},
  year={2021},
  publisher={Nature Publishing Group UK London}
}

@article{martin2021implicit,
  title={Implicit self-regularization in deep neural networks: Evidence from random matrix theory and implications for learning},
  author={Martin, Charles H and Mahoney, Michael W},
  journal={JMLR},
  year={2021}
}

@article{dong2023raft,
  title={Raft: Reward ranked finetuning for generative foundation model alignment},
  author={Dong, Hanze and Xiong, Wei and Goyal, Deepanshu and Zhang, Yihan and Chow, Winnie and Pan, Rui and Diao, Shizhe and Zhang, Jipeng and Shum, Kashun and Zhang, Tong},
  journal={arXiv preprint arXiv:2304.06767},
  year={2023}
}

@inproceedings{zheng2024llamafactory,
  title={Llamafactory: Unified efficient fine-tuning of 100+ language models},
  author={Zheng, Yaowei and Zhang, Richong and Zhang, Junhao and Ye, Yanhan and Luo, Zheyan},
  booktitle={ACL},
  year={2024}
}

@article{huan2025does,
  title={Does math reasoning improve general llm capabilities? understanding transferability of llm reasoning},
  author={Huan, Maggie and Li, Yuetai and Zheng, Tuney and Xu, Xiaoyu and Kim, Seungone and Du, Minxin and Poovendran, Radha and Neubig, Graham and Yue, Xiang},
  journal={arXiv preprint arXiv:2507.00432},
  year={2025}
}
\bibliographystyle{colm2026_conference}

\newpage
\clearpage\newpage
\onecolumn
\section*{\Large{Appendix}}
\setcounter{section}{0}
\setcounter{figure}{0}
\setcounter{table}{0}
\makeatletter 
\renewcommand{\thesection}{\Alph{section}}
\renewcommand{\theHsection}{\Alph{section}}
\renewcommand{\thefigure}{A\arabic{figure}} 
\renewcommand{\theHfigure}{A\arabic{figure}} 
\renewcommand{\thetable}{A\arabic{table}}
\renewcommand{\theHtable}{A\arabic{table}}
\makeatother
\renewcommand{\thetable}{A\arabic{table}}
\setcounter{mylemma}{0}
\renewcommand{\themylemma}{A\arabic{mylemma}}
\setcounter{equation}{0}
\renewcommand{\theequation}{A\arabic{equation}}
\setcounter{algorithm}{0}
\renewcommand{\thealgorithm}{A\arabic{algorithm}}
\renewcommand{\theHalgorithm}{A\arabic{algorithm}} 

\section{Additional Visualizations of Spectral Signals}
\label{appen:spec-sig}

\begin{figure*}[h]
  \centering
  \vspace{-0mm}
  \begin{tabular}{@{}c@{\hspace{4mm}}c@{}}
    \includegraphics[width=0.4\linewidth]{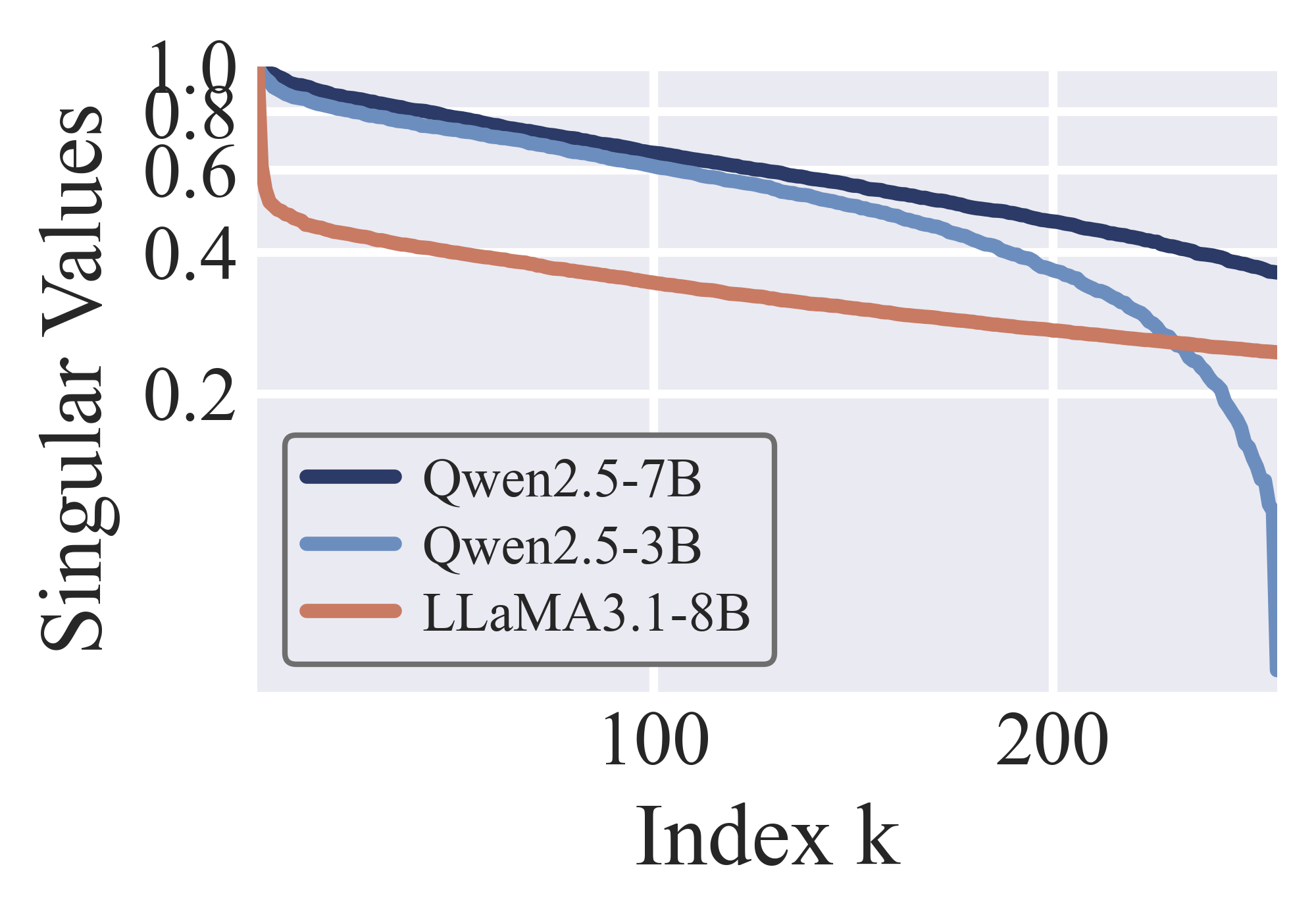} &
    \includegraphics[width=0.4\linewidth]{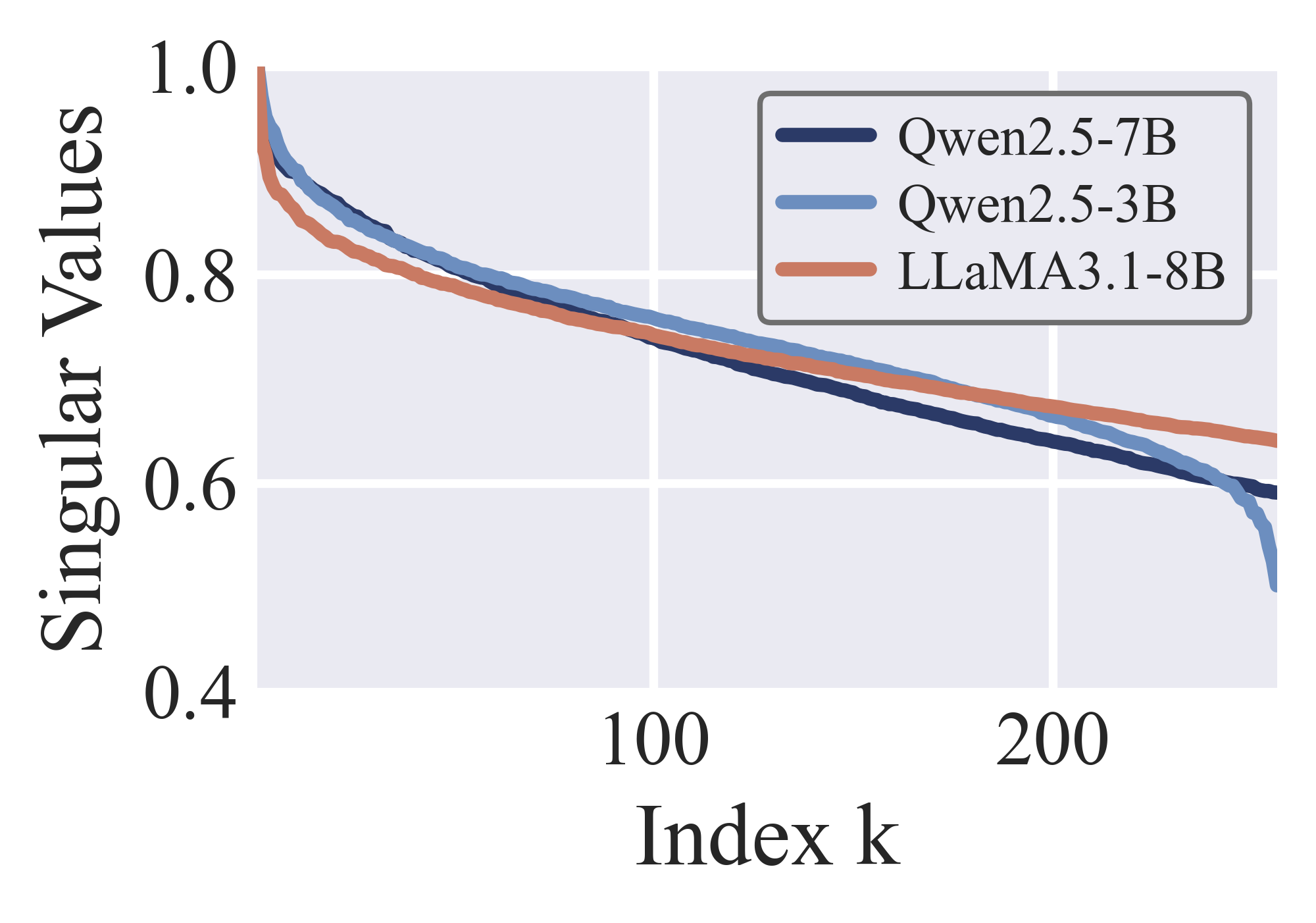} \\
    \small (a) $K$ projection &
    \small (b) $V$ projection \\[0mm]
    
    \includegraphics[width=0.4\linewidth]{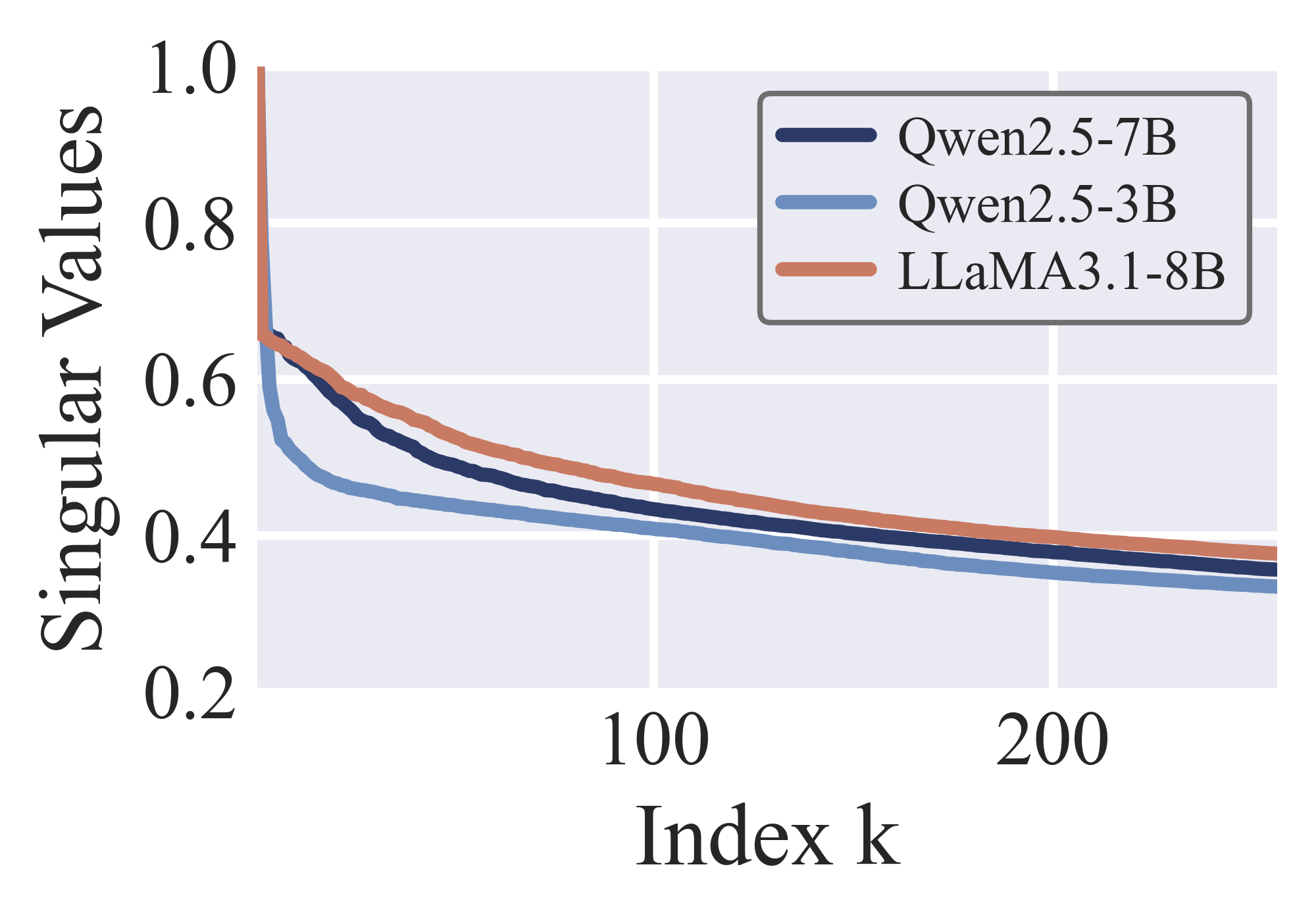} &
    \includegraphics[width=0.4\linewidth]{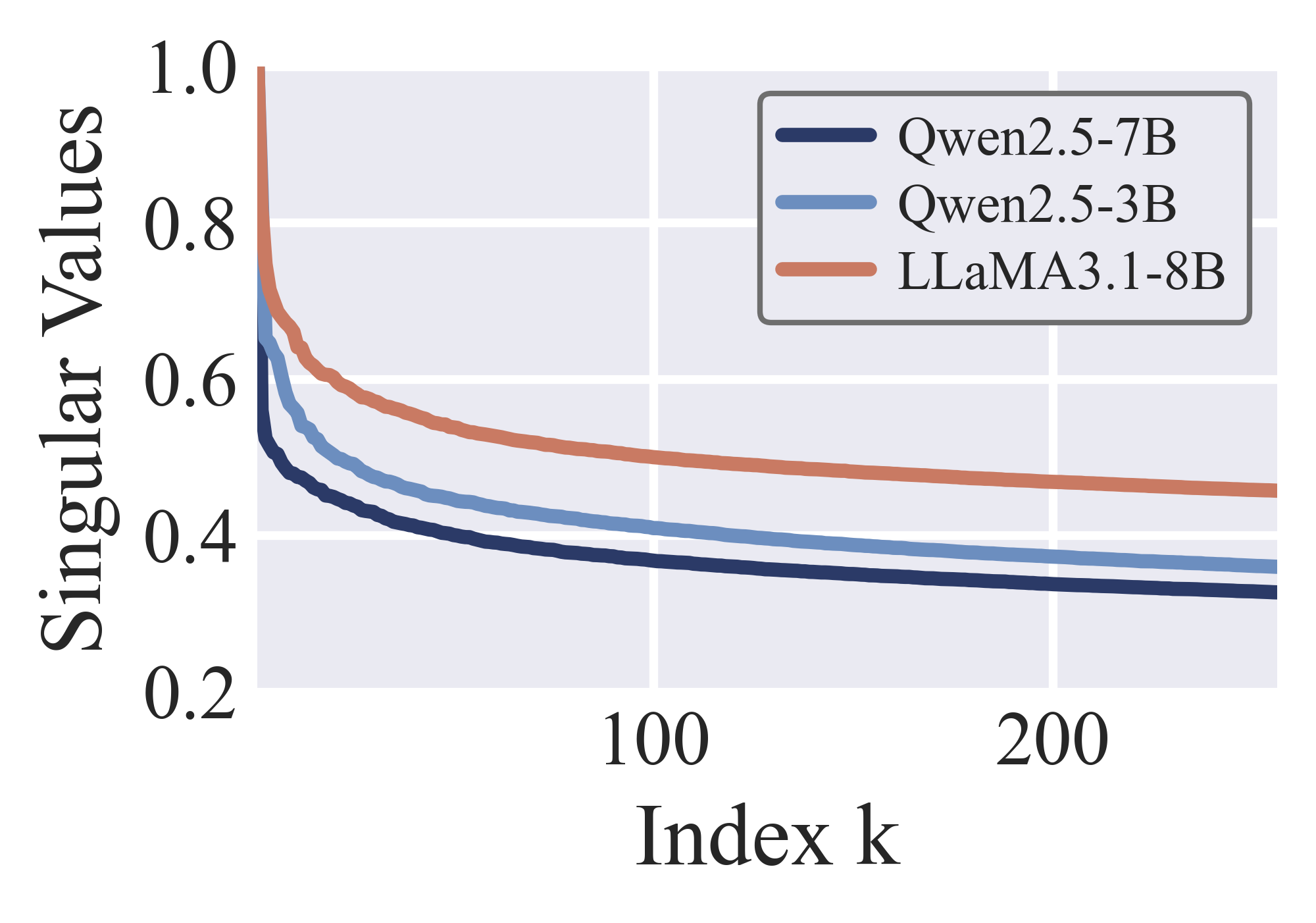} \\
    \small (c) $O$ projection &
    \small (d) $\text{FFN}_{\text{DOWN}}$ projection\\
  \end{tabular}
  \vspace{0mm}
  \caption{\small
Singular value distributions of middle-layer weight matrices for component types $c \in \{K, V, O, \text{FFN}_{\text{DOWN}}\}$ across models.
Experiment setup is consistent with Fig.\,\ref{fig:svd-distribution}.
  }
  \vspace{0mm}
  \label{fig:svd-distribution-appendix}
\end{figure*}

\begin{figure*}[h]
  \centering
  \vspace{0mm}
  \begin{tabular}{@{}c@{\hspace{4mm}}c@{}}
    \includegraphics[width=0.4\linewidth]{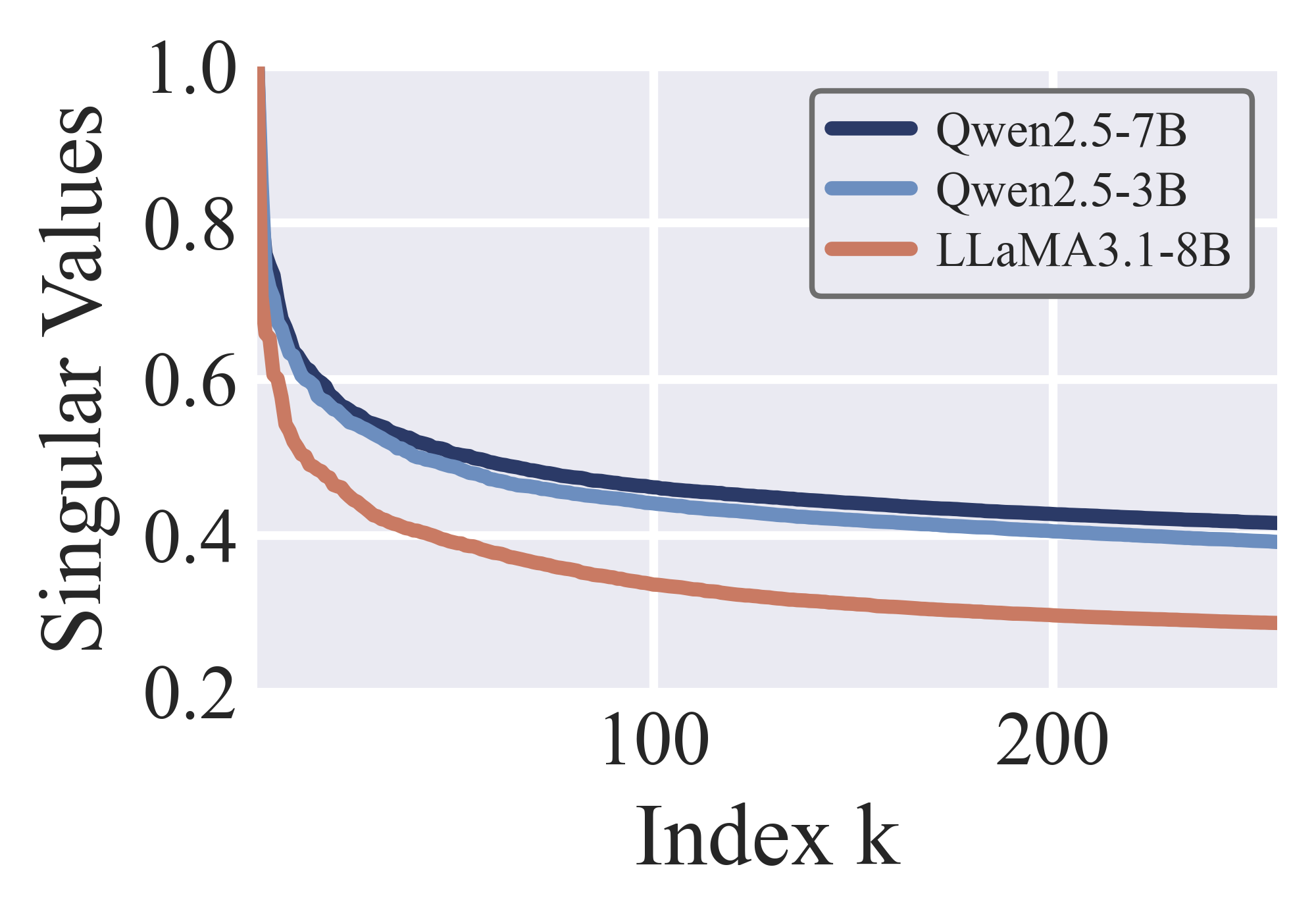} &
    \includegraphics[width=0.4\linewidth]{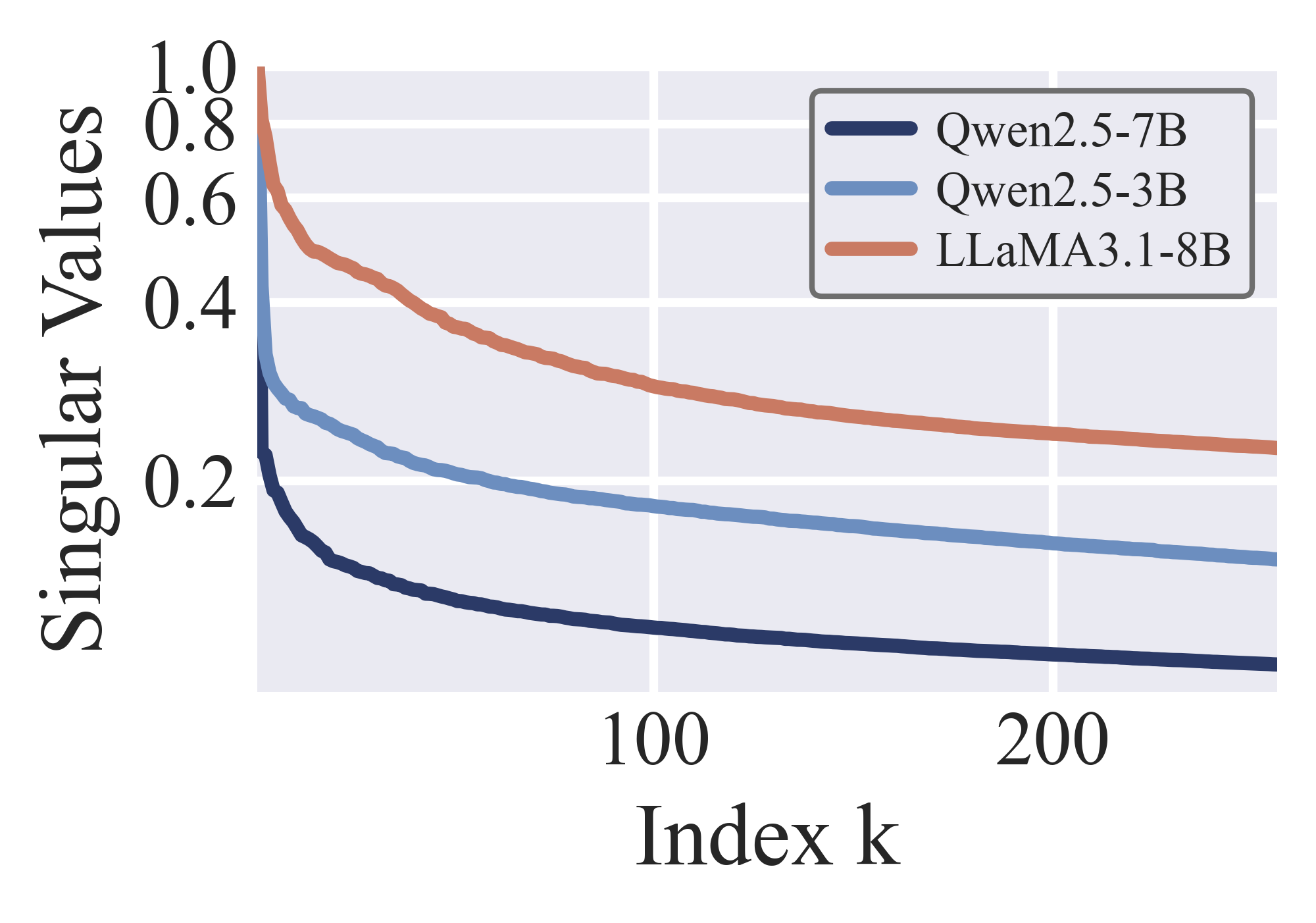} \\
    \small (a) $\text{FFN}_{\text{DOWN}}$ &
    \small (b) $\text{FFN}_{\text{GATE}}$ \\[2mm]
    
    \includegraphics[width=0.4\linewidth]{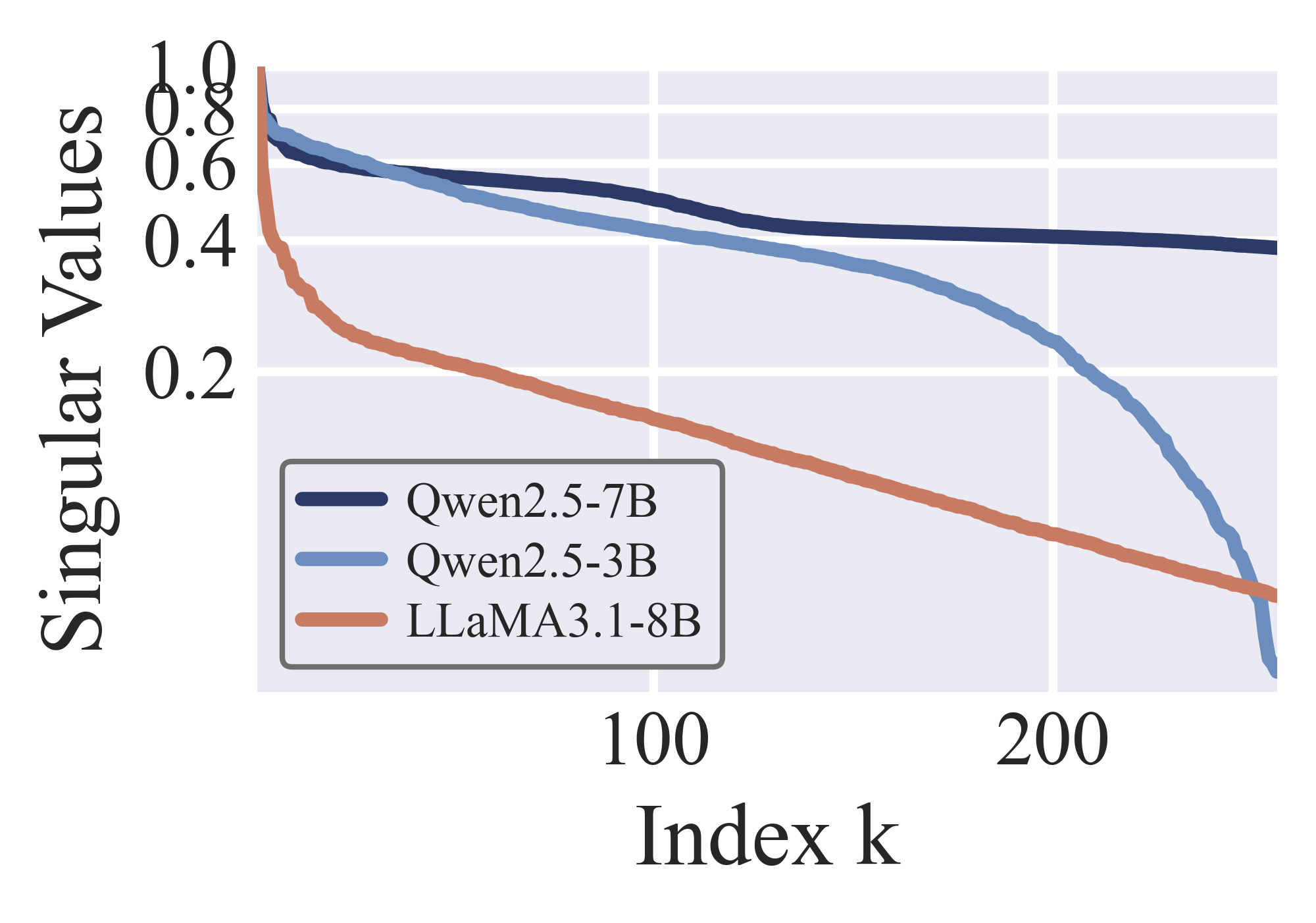} &
    \includegraphics[width=0.4\linewidth]{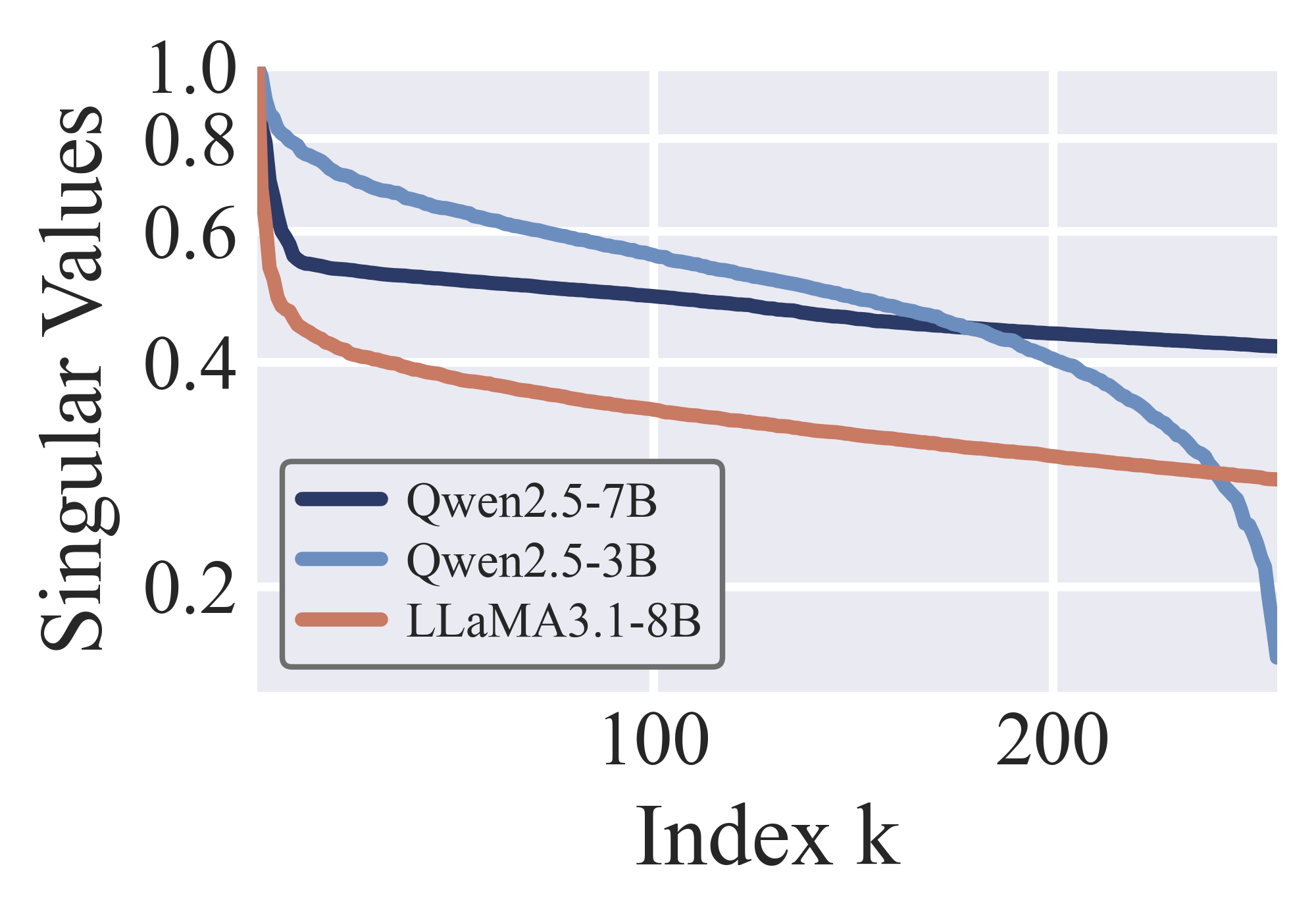} \\
    \small (c) $K$ projection &
    \small (d) $V$ projection \\
  \end{tabular}
  \vspace{0mm}
  \caption{\small
Singular value distributions of \textit{first-layer} weight matrices for component types $c \in \{\text{FFN}_{\text{DOWN}}, \text{FFN}_{\text{GATE}}, K, V\}$ across models.
The other experiment setup is consistent with Fig.\,\ref{fig:svd-distribution}.
  }
  \vspace{0mm}
  \label{fig:svd-distribution-first-layer}
\end{figure*}

\clearpage
\section{Model Zoo of Each Regime}
\label{appen:model-zoo}

This section catalogs the model pairs used in our experiments, organized by the three lineage discrimination regimes defined in Sec.\,\ref{sec:problem}: independent-origin (Scenario 1), same-series (Scenario 2), and shared-base (Scenario 3).
We reserve the same-series regime for scale variants within the same model generation, such as Qwen2.5 models of different sizes or Qwen3 models of different sizes.
Cross-generation pairs between Qwen2.5 and Qwen3 are instead treated as intermediate cross-series relationships and analyzed separately in Appendix~\ref{app:cross-series}.
These pairs form the empirical basis for evaluating coarse-grained regime separation and fine-grained discrimination capabilities.

\subsection{Independent-Origin Regime (Scenario 1)}

Model pairs in this regime originate from different organizations or training pipelines, with no shared base model.
The complete set comprises 27 model pairs across diverse model families:

\textbf{Qwen cross-family pairs (7 pairs)}:
\begin{itemize}[leftmargin=*, itemsep=1pt, parsep=0pt]
    \item Qwen/Qwen2.5-7B vs. meta-llama/Llama-2-7b-hf.
    \item Qwen/Qwen2.5-7B vs. meta-llama/Llama-3.1-8B.
    \item Qwen/Qwen2.5-7B vs. mistralai/Mistral-7B-v0.3.
    \item Qwen/Qwen2.5-7B vs. google/gemma-7b.
    \item Qwen/Qwen2.5-7B vs. 01-ai/Yi-6B.
    \item Qwen/Qwen2.5-7B vs. tiiuae/falcon-7b.
    \item Qwen/Qwen2.5-7B vs. deepseek-ai/deepseek-llm-7b-base.
\end{itemize}

\textbf{LLaMA cross-family pairs (10 pairs)}:
\begin{itemize}[leftmargin=*, itemsep=1pt, parsep=0pt]
    \item meta-llama/Llama-2-7b-hf vs. google/gemma-7b.
    \item meta-llama/Llama-2-7b-hf vs. mistralai/Mistral-7B-v0.3.
    \item meta-llama/Llama-2-7b-hf vs. 01-ai/Yi-6B.
    \item meta-llama/Llama-2-7b-hf vs. tiiuae/falcon-7b.
    \item meta-llama/Llama-2-7b-hf vs. deepseek-ai/deepseek-llm-7b-base.
    \item meta-llama/Llama-3.1-8B vs. mistralai/Mistral-7B-v0.3.
    \item meta-llama/Llama-3.1-8B vs. google/gemma-7b.
    \item meta-llama/Llama-3.1-8B vs. 01-ai/Yi-6B.
    \item meta-llama/Llama-3.1-8B vs. tiiuae/falcon-7b.
    \item meta-llama/Llama-3.1-8B vs. deepseek-ai/deepseek-llm-7b-base.
\end{itemize}

\textbf{Other cross-family pairs (10 pairs)}:
\begin{itemize}[leftmargin=*, itemsep=1pt, parsep=0pt]
    \item mistralai/Mistral-7B-v0.3 vs. google/gemma-7b.
    \item mistralai/Mistral-7B-v0.3 vs. 01-ai/Yi-6B.
    \item mistralai/Mistral-7B-v0.3 vs. tiiuae/falcon-7b.
    \item mistralai/Mistral-7B-v0.3 vs. deepseek-ai/deepseek-llm-7b-base.
    \item google/gemma-7b vs. 01-ai/Yi-6B.
    \item google/gemma-7b vs. tiiuae/falcon-7b.
    \item google/gemma-7b vs. deepseek-ai/deepseek-llm-7b-base.
    \item 01-ai/Yi-6B vs. tiiuae/falcon-7b.
    \item 01-ai/Yi-6B vs. deepseek-ai/deepseek-llm-7b-base.
    \item tiiuae/falcon-7b vs. deepseek-ai/deepseek-llm-7b-base.
\end{itemize}

These 27 pairs represent the most distinct lineage relationships, spanning organizations including Alibaba (Qwen2.5), Meta (LLaMA), Mistral AI, Google (Gemma), 01.AI (Yi), TII (Falcon), and DeepSeek.
Models differ in architecture details, training data composition, vocabulary, and tokenization schemes, producing maximally divergent weight-space structures.

\subsection{Same-Series Regime (Scenario 2)}

Model pairs in this regime belong to the same model series and share architectures and training pipelines, but differ primarily in parameter scale.
The set comprises 10 model pairs:

\textbf{Qwen series (4 pairs)}:
\begin{itemize}[leftmargin=*, itemsep=1pt, parsep=0pt]
    \item Qwen/Qwen2.5-3B vs. Qwen/Qwen2.5-7B.
    \item Qwen/Qwen2.5-7B vs. Qwen/Qwen2.5-14B.
    \item Qwen/Qwen3-4B vs. Qwen/Qwen3-8B.
    \item Qwen/Qwen3-8B vs. Qwen/Qwen3-14B.
\end{itemize}

\textbf{Gemma series (1 pair)}:
\begin{itemize}[leftmargin=*, itemsep=1pt, parsep=0pt]
    \item google/gemma-2b vs. google/gemma-7b.
\end{itemize}

\textbf{Pythia series (2 pairs)}:
\begin{itemize}[leftmargin=*, itemsep=1pt, parsep=0pt]
    \item EleutherAI/pythia-1b vs. EleutherAI/pythia-1.4b.
    \item EleutherAI/pythia-1.4b vs. EleutherAI/pythia-2.8b.
\end{itemize}

\textbf{Falcon series (1 pair)}:
\begin{itemize}[leftmargin=*, itemsep=1pt, parsep=0pt]
    \item tiiuae/falcon-7b vs. tiiuae/falcon-11b.
\end{itemize}

\textbf{Mistral/Mixtral series (1 pair)}:
\begin{itemize}[leftmargin=*, itemsep=1pt, parsep=0pt]
    \item mistralai/Mistral-7B-v0.3 vs. mistralai/Mixtral-8x7B-v0.1.
\end{itemize}

\textbf{LLaMA-2 series (1 pair)}:
\begin{itemize}[leftmargin=*, itemsep=1pt, parsep=0pt]
    \item meta-llama/Llama-2-7b-hf vs. meta-llama/Llama-2-13b-hf.
\end{itemize}

These pairs share training methodology and architectural design but exhibit scale-induced variations in weight space structure.
Scale ratios range from 1.4$\times$ (Pythia-1B vs. 1.4B) to 8$\times$ (Mistral-7B vs. Mixtral-8x7B with MoE architecture).

\subsection{Shared-Base Regime (Scenario 3)}

Model pairs in this regime originate from a common pretrained base model and diverge through post-training procedures.
We distinguish two subcategories based on fine-grained variation sources.

\paragraph{Data-Scale Variants.}

These pairs share the same base model and post-training algorithm but differ in the size of the fine-tuning dataset:

\textbf{Standard instruction tuning (5 pairs)}:
\begin{itemize}[leftmargin=*, itemsep=1pt, parsep=0pt]
    \item Qwen/Qwen2.5-7B vs. Qwen/Qwen2.5-7B-Instruct.
    \item Qwen/Qwen2.5-14B vs. Qwen/Qwen2.5-14B-Instruct.
    \item Qwen/Qwen3-4B vs. Qwen/Qwen3-4B-Instruct.
    \item Qwen/Qwen3-4B vs. Qwen/Qwen3-4B-Thinking.
    \item meta-llama/Llama-3.1-8B vs. meta-llama/Meta-Llama-3.1-8B-Instruct.
\end{itemize}

\textbf{Controlled data-scale ablation (4 pairs, meta-llama/Llama-3.1-8B base)}:
\begin{itemize}[leftmargin=*, itemsep=1pt, parsep=0pt]
    \item meta-llama/Llama-3.1-8B vs. sft\_alpaca\_10pct\_lr1e5 (10\% of Alpaca dataset).
    \item meta-llama/Llama-3.1-8B vs. sft\_alpaca\_25pct\_lr1e5 (25\% of Alpaca dataset).
    \item meta-llama/Llama-3.1-8B vs. sft\_alpaca\_50pct\_lr1e5 (50\% of Alpaca dataset).
    \item meta-llama/Llama-3.1-8B vs. sft\_alpaca\_100pct\_lr1e5 (100\% of Alpaca dataset).
\end{itemize}

The controlled ablation series enables systematic evaluation of how dataset scale affects weight-space geometry while holding all other factors constant.

\paragraph{Algorithmic Variants.}

These pairs share the same base model but differ in the post-training algorithm applied, isolating the effect of algorithmic choice on weight structure:

\textbf{meta-llama/Llama-3.1-8B with Tulu-3 variants (3 pairs)}:
\begin{itemize}[leftmargin=*, itemsep=1pt, parsep=0pt]
    \item meta-llama/Llama-3.1-8B vs. allenai/Llama-3.1-Tulu-3-8B.
    \item meta-llama/Llama-3.1-8B vs. allenai/Llama-3.1-Tulu-3-8B-SFT.
    \item meta-llama/Llama-3.1-8B vs. allenai/Llama-3.1-Tulu-3-8B-DPO.
\end{itemize}

\textbf{Qwen/Qwen2.5-Math-7B with RLHFlow variants (5 pairs)}:
\begin{itemize}[leftmargin=*, itemsep=1pt, parsep=0pt]
    \item Qwen/Qwen2.5-Math-7B vs. RLHFlow/Qwen2.5-7B-SFT.
    \item Qwen/Qwen2.5-Math-7B vs. RLHFlow/Qwen2.5-7B-DPO.
    \item Qwen/Qwen2.5-Math-7B vs. RLHFlow/Qwen2.5-7B-DPO-Zero.
    \item Qwen/Qwen2.5-Math-7B vs. RLHFlow/Qwen2.5-7B-PPO-Zero.
    \item Qwen/Qwen2.5-Math-7B vs. RLHFlow/Qwen2.5-7B-RAFT-Zero.
\end{itemize}

\textbf{Qwen3 with reasoning-specialized variants (3 pairs)} \citep{huan2025does}:
\begin{itemize}[leftmargin=*, itemsep=1pt, parsep=0pt]
    \item Qwen/Qwen3-14B vs. ReasoningTransferability/UniReason-Qwen3-14B-no-think-SFT.
    \item Qwen/Qwen3-14B vs. ReasoningTransferability/UniReason-Qwen3-14B-think-SFT.
    \item Qwen/Qwen3-14B vs. ReasoningTransferability/UniReason-Qwen3-14B-RL. 
\end{itemize}

These algorithmic variants span supervised fine-tuning (SFT), direct preference optimization (DPO), proximal policy optimization (PPO), and rejection sampling fine-tuning (RAFT), enabling systematic comparison of how different alignment algorithms induce geometric transformations in weight space.

\textbf{Gaussian Noise Augmentation for Controlled Experiments.}
To systematically evaluate robustness and expand the experimental coverage, we augment the model zoo with synthetic variants created via controlled Gaussian noise perturbation.
For each model in the dataset, we create a noise-perturbed copy by adding Gaussian noise $\mathcal{N}(0, \sigma^2 I)$ to all weight matrices, where $\sigma = 0.0001$ is the noise standard deviation.
This yields an additional set of model pairs with the same cardinality as the original dataset, effectively doubling the total number of pairs across all three regimes.

The noise magnitude $\sigma = 0.0001$ is chosen to be small enough to preserve model functionality (general utility) while introducing measurable weight-space perturbations.

\clearpage
\section{Spectral Magnitude-based Fingerprint Algorithm}
\label{appen:trace-algo}

\begin{algorithm}[h]
\caption{Spectral Trace Fingerprint and Comparison}
\label{alg:trace-fingerprint}
\small
\begin{algorithmic}[1]

\Require Models $\mathcal{M}_a, \mathcal{M}_b$; components list $\mathcal{C}$
\Ensure Similarity score $S(\mathcal{M}_a, \mathcal{M}_b)$
\For{each component $c \in \mathcal{C}$}

    \Function{ExtractTrace}{$\mathcal{M}, c$}
        \State $\tau_c \gets \emptyset$
        \For{each layer $l$}
            \State $W \gets W_c^{(l)}$
            \State $t \gets \sqrt{\mathrm{tr}(W^\top W)}$ \Comment{total spectral energy}
            \State append $t$ to $\tau_c$
        \EndFor
        \State \Return $\tau_c$
    \EndFunction

    \State $\tau_c^a \gets \textsc{ExtractTrace}(\mathcal{M}_a, c)$
    \State $\tau_c^b \gets \textsc{ExtractTrace}(\mathcal{M}_b, c)$

    \State $L \gets \max(|\tau_c^a|, |\tau_c^b|)$
    \State $\tilde{\tau}_c^a, \tilde{\tau}_c^b \gets \textsc{Interp}(\tau_c^a, \tau_c^b \rightarrow L)$

    \State $S_c \gets \text{PearsonCorr}(\tilde{\tau}_c^a, \tilde{\tau}_c^b)$
\EndFor
\State $S \gets \frac{1}{|\mathcal{C}|} \sum_{c} S_c$
\State \Return $S$
\end{algorithmic}
\end{algorithm}

\section{Additional Settings and Results of Spectral Energy}
\label{appen:spectral-energy}

\begin{table*}[htb]
\centering
\small
\scriptsize
\setlength{\tabcolsep}{5pt}
\caption{\small
Similarity scores for Scenario 1 (independent-origin) before and after Gaussian perturbation ($\sigma^2=10^{-4}$) under baselines and our Trace-based similarity score.}
\vspace{2mm}
\begin{tabular}{lcccc|cccc}
\toprule
\multirow{2}{*}{\textbf{Model Pair}} 
& \multicolumn{4}{c}{\textbf{Original}} 
& \multicolumn{4}{c}{\textbf{Noise}} \\
\cmidrule(lr){2-5} \cmidrule(lr){6-9}
& AWM & HuReF & PDF & Trace 
& AWM & HuReF & PDF & Trace \\
\midrule

Qwen2.5-7B vs LLaMA-2-7B & 0.0023 & -0.066 & 0.7453 & 0.3082 & 0.0023 & -0.066 & 0.745 & 0.3077 \\
Qwen2.5-7B vs LLaMA-3.1-8B & 0.0028 & 0.0092 & 0.9159 & 0.3522 & 0.0028 & 0.0092 & 0.916 & 0.352 \\
Qwen2.5-7B vs Mistral-7B-v0.3 & 0.0019 & 0.0282 & 0.9313 & 0.2261 & 0.0019 & 0.0282 & 0.9312 & 0.227 \\
Qwen2.5-7B vs Gemma-7B & 0.0019 & -0.0244 & 0.4919 & 0.0321 & 0.0018 & -0.0244 & 0.492 & 0.1736 \\

LLaMA-3.1-8B vs Mistral-7B-v0.3 & 0.0056 & -0.0186 & 0.9516 & 0.6222 & 0.0055 & -0.0186 & 0.9517 & 0.6221 \\
LLaMA-2-7B vs Gemma-7B & 0.0024 & -0.0977 & 0.6029 & 0.2507 & 0.0024 & -0.0977 & 0.6029 & 0.1412 \\
LLaMA-2-7B vs Mistral-7B-v0.3 & 0.0067 & 0.0277 & 0.8594 & 0.2568 & 0.0067 & 0.0277 & 0.8594 & 0.2572 \\
LLaMA-3.1-8B vs Gemma-7B & 0.0027 & 0.0358 & 0.5225 & 0.2977 & 0.0027 & 0.0358 & 0.5223 & 0.1035 \\

Mistral-7B-v0.3 vs Gemma-7B & 0.0026 & -0.0286 & 0.4819 & 0.2507 & 0.0026 & -0.0286 & 0.4818 & 0.1174 \\

Yi-6B vs LLaMA-2-7B & 0.0067 & 0.0663 & 0.8903 & 0.3648 & 0.0067 & 0.0663 & 0.8904 & 0.3648 \\
Yi-6B vs Qwen2.5-7B & 0.0020 & -0.0474 & 0.8381 & 0.5845 & 0.0019 & -0.0474 & 0.8381 & 0.5838 \\

Falcon-7B vs LLaMA-2-7B & 0.0005 & -0.0097 & 0.5393 & 0.3396 & 0.0005 & -0.0097 & 0.5394 & 0.3396 \\
Falcon-7B vs Qwen2.5-7B & 0.0008 & -0.0108 & 0.6956 & 0.2753 & 0.0008 & -0.0108 & 0.6954 & 0.2753 \\

DeepSeek-LLM-7B vs LLaMA-3.1-8B & 0.0084 & -0.0269 & 0.8899 & 0.2490 & 0.0084 & -0.0269 & 0.890 & 0.2486 \\
DeepSeek-LLM-7B vs Mistral-7B-v0.3 & 0.0054 & 0.0810 & 0.9276 & 0.1696 & 0.0053 & 0.0810 & 0.9276 & 0.1711 \\

Qwen2.5-7B vs DeepSeek-LLM-7B & 0.0027 & 0.0818 & 0.8301 & 0.2419 & 0.0027 & 0.0818 & 0.8299 & 0.2417 \\
LLaMA-2-7B vs DeepSeek-LLM-7B & 0.0083 & 0.3846 & 0.9562 & 0.3595 & 0.0083 & 0.3846 & 0.9562 & 0.3599 \\

LLaMA-3.1-8B vs Yi-6B & 0.0046 & 0.0178 & 0.8523 & 0.5080 & 0.0046 & 0.0178 & 0.8521 & 0.5084 \\
LLaMA-3.1-8B vs Falcon-7B & 0.0003 & 0.0047 & 0.6745 & 0.4542 & 0.0003 & 0.0047 & 0.6743 & 0.4542 \\

Mistral-7B-v0.3 vs Yi-6B & 0.0059 & -0.0348 & 0.8896 & 0.3506 & 0.0059 & -0.0348 & 0.8895 & 0.3501 \\
Mistral-7B-v0.3 vs Falcon-7B & 0.0004 & -0.0026 & 0.6139 & 0.3030 & 0.0003 & -0.0026 & 0.6137 & 0.3030 \\

Gemma-7B vs Yi-6B & 0.0031 & 0.0872 & 0.5348 & 0.1951 & 0.0031 & 0.0872 & 0.5348 & 0.0717 \\
Gemma-7B vs Falcon-7B & 0.0010 & 0.0034 & 0.7746 & 0.1472 & 0.0010 & 0.0034 & 0.7747 & 0.0842 \\
Gemma-7B vs DeepSeek-LLM-7B & 0.0025 & -0.0321 & 0.5369 & 0.1337 & 0.0025 & -0.0321 & 0.5369 & 0.1496 \\

Yi-6B vs Falcon-7B & 0.0004 & 0.0034 & 0.5698 & 0.4126 & 0.0004 & 0.0034 & 0.5696 & 0.4126 \\
Yi-6B vs DeepSeek-LLM-7B & 0.0050 & -0.0136 & 0.9328 & 0.4036 & 0.0050 & -0.0136 & 0.9328 & 0.4039 \\

Falcon-7B vs DeepSeek-LLM-7B & 0.0011 & 0.0004 & 0.5400 & 0.3018 & 0.0011 & 0.0004 & 0.5399 & 0.3018 \\

\bottomrule
\vspace{4mm}
\end{tabular}
\end{table*}

\begin{table}[t]
\centering
\small
\scriptsize
\vspace{2mm}
\caption{\small
Similarity scores for Scenario 2 (same-series) before and after Gaussian perturbation ($\sigma^2=10^{-4}$) under baselines and our Trace-based similarity scores.}
\vspace{2mm}
\setlength{\tabcolsep}{4pt}
\begin{tabular}{lcccc|cccc}
\toprule
\multirow{2}{*}{\textbf{Model Pair}} 
& \multicolumn{4}{c}{\textbf{Original}} 
& \multicolumn{4}{c}{\textbf{Noise}} \\
\cmidrule(lr){2-5} \cmidrule(lr){6-9}
& AWM & HuReF & PDF & Trace 
& AWM & HuReF & PDF & Trace \\
\midrule

Qwen2.5-3B vs Qwen2.5-7B & 0.0020 & NA & 0.9440 & 0.5738 & 0.0020 & NA & 0.9440 & 0.5741 \\
Qwen2.5-7B vs Qwen2.5-14B & 0.0034 & NA & 0.8133 & 0.5220 & 0.0033 & NA & 0.8133 & 0.5219 \\
Qwen3-4B vs Qwen3-8B & 0.0842 & NA & 0.9567 & 0.6904 & 0.0842 & NA & 0.9567 & 0.6904 \\
Qwen3-8B vs Qwen3-14B & 0.0134 & NA & 0.8502 & 0.6109 & 0.0134 & NA & 0.8502 & 0.6109 \\
Gemma-2B vs Gemma-7B & 0.0021 & NA & 0.5662 & 0.6169 & 0.0021 & NA & 0.5662 & 0.0298 \\
Pythia-1B vs Pythia-1.4B & 0.0202 & NA & 0.5830 & 0.6979 & 0.0202 & NA & 0.5830 & 0.6979 \\
Pythia-1.4B vs Pythia-2.8B & 0.0045 & NA & 0.9136 & 0.9298 & 0.0046 & NA & 0.9136 & 0.6465 \\
Falcon-7B vs Falcon-11B & 0.0007 & NA & 0.6654 & 0.5368 & 0.0007 & NA & 0.6654 & 0.5368 \\
Mistral-7B-v0.3 vs Mixtral-8$\times$7B & 0.6665 & NA & 0.9856 & 0.8679 & 0.6664 & NA & 0.9856 & 0.4265 \\
LLaMA-2-7B vs LLaMA-2-13B & 0.0118 & NA & 0.9840 & 0.4512 & 0.0119 & NA & 0.9840 & 0.4512 \\

\bottomrule
\vspace{2mm}
\end{tabular}
\end{table}

\begin{table*}[t]
\vspace{2mm}
\centering
\scriptsize
\caption{\small
Similarity scores for Scenario 3 (shared-base) before and after Gaussian perturbation ($\sigma^2=10^{-4}$) under baselines and our Trace-based similarity scores.}
\vspace{2mm}
\setlength{\tabcolsep}{3pt}
\begin{tabular}{lcccc|cccc}
\toprule
\multirow{2}{*}{\textbf{Model Pair}} 
& \multicolumn{4}{c}{\textbf{Original}} 
& \multicolumn{4}{c}{\textbf{Noise}} \\
\cmidrule(lr){2-5} \cmidrule(lr){6-9}
& AWM & HuReF & PDF & Trace 
& AWM & HuReF & PDF & Trace \\
\midrule

\multicolumn{9}{c}{\textbf{Data-scale variants}} \\
\midrule

Qwen2.5-7B vs Qwen2.5-7B-Instruct & 0.999 & 0.999 & 1.000 & 0.998 & 0.999 & 0.999 & 1.000 & 0.999 \\
Qwen2.5-14B vs Qwen2.5-14B-Instruct & 0.999 & 1.000 & 1.000 & 0.998 & 0.999 & 1.000 & 1.000 & 0.999 \\
Qwen3-4B vs Qwen3-4B-Instruct & 0.998 & 0.999 & 0.999 & 0.999 & 0.999 & 0.999 & 0.999 & 0.999 \\
Qwen3-4B vs Qwen3-4B-Thinking & 1.000 & 0.999 & 0.998 & 0.999 & 1.000 & 0.999 & 0.998 & 0.999 \\

LLaMA-3.1-8B vs LLaMA-3.1-8B-Instruct & 0.998 & 0.997 & 0.999 & 0.997 & 0.999 & 0.998 & 0.999 & 0.998 \\
LLaMA-3.1-8B vs Alpaca-10\% & 0.999 & 0.998 & 1.000 & 1.000 & 0.999 & 0.999 & 1.000 & 1.000 \\
LLaMA-3.1-8B vs Alpaca-25\% & 0.999 & 0.999 & 0.999 & 0.998 & 0.999 & 0.999 & 0.999 & 0.999 \\
LLaMA-3.1-8B vs Alpaca-50\% & 1.000 & 1.000 & 1.000 & 0.998 & 1.000 & 1.000 & 1.000 & 0.999 \\
LLaMA-3.1-8B vs Alpaca-100\% & 1.000 & 1.000 & 1.000 & 0.998 & 1.000 & 1.000 & 1.000 & 0.999 \\

\midrule
\multicolumn{9}{c}{\textbf{Algorithmic variants}} \\
\midrule

LLaMA-3.1-8B vs Tulu-3-8B & 1.000 & 1.000 & 1.000 & 0.999 & 1.000 & 1.000 & 1.000 & 0.999 \\
LLaMA-3.1-8B vs Tulu-3-8B-SFT & 0.999 & 0.999 & 1.000 & 0.998 & 0.999 & 0.999 & 1.000 & 0.999 \\
LLaMA-3.1-8B vs Tulu-3-8B-DPO & 0.998 & 0.999 & 0.999 & 0.999 & 0.999 & 0.999 & 0.999 & 1.000 \\

Qwen2.5-Math-7B vs Qwen2.5-7B-SFT & 1.000 & 1.000 & 0.999 & 0.999 & 1.000 & 1.000 & 0.999 & 0.999 \\
Qwen2.5-Math-7B vs Qwen2.5-7B-DPO & 0.998 & 0.996 & 0.998 & 0.998 & 0.999 & 0.998 & 0.999 & 0.999 \\
Qwen2.5-Math-7B vs Qwen2.5-7B-DPO-Zero & 1.000 & 1.000 & 0.999 & 0.999 & 1.000 & 1.000 & 0.999 & 1.000 \\
Qwen2.5-Math-7B vs Qwen2.5-7B-PPO-Zero & 0.997 & 0.998 & 0.997 & 0.999 & 0.998 & 0.999 & 0.998 & 1.000 \\
Qwen2.5-Math-7B vs Qwen2.5-7B-RAFT-Zero & 1.000 & 1.000 & 0.999 & 0.999 & 1.000 & 1.000 & 0.999 & 1.000 \\

Qwen3-14B vs UniReason-no-think & 0.998 & 0.999 & 0.998 & 0.999 & 0.999 & 0.999 & 0.999 & 0.999 \\
Qwen3-14B vs UniReason-think & 1.000 & 0.999 & 1.000 & 0.999 & 1.000 & 0.999 & 1.000 & 1.000 \\
Qwen3-14B vs UniReason-RL & 0.998 & 0.999 & 0.999 & 0.997 & 0.999 & 0.999 & 0.999 & 0.998 \\

\bottomrule
\vspace{2mm}
\end{tabular}
\label{tab:scenario3}
\end{table*}

In Scenario 1 (independent-origin), models exhibit substantial structural differences in weight space, where AWM and HuReF concentrate near zero and PDF saturates at high values, while Trace produces a wide range of scores that clearly separates model families and remain stable under perturbation. 
In Scenario 2 (same-series), similarity increases due to shared architecture and training pipeline, with PDF showing partial saturation, whereas the Trace-based discrimination method consistent variations by model scale and preserves these differences under noise. 
In Scenario 3 (shared-base), all methods yield near-saturated scores close to 1 due to shared initialization; however, Trace still reveals small but consistent deviations across both data-scale and algorithmic variants, capturing subtle directional changes that remain robust under perturbation.

\clearpage

\section{Discrimination across Adjacent Lineage Regimes}
\label{app:adjacent-regimes}

We further examine discrimination across both adjacent regime boundaries.
Distinguishing independent-origin (S1) from same-series (S2) models is particularly challenging because models within the same series may differ substantially in scale while retaining related spectral structure.
By contrast, shared-base (S3) models inherit the same pretrained checkpoint and are therefore expected to exhibit substantially stronger similarity.

\begin{table}[!ht]
\centering
\vspace{1mm}
\caption{\small
AUC scores for distinguishing adjacent lineage regimes.
S1, S2, and S3 denote independent-origin, same-series, and shared-base,
respectively.}
\label{tab:adjacent-auc}
\vspace{1mm}
\small
\renewcommand{\arraystretch}{1.08}
\begin{tabular*}{0.58\textwidth}{@{\extracolsep{\fill}}lcc}
\toprule
\textbf{Method}
& \textbf{S1 vs.\ S2}
& \textbf{S2 vs.\ S3} \\
\midrule
AWM   & 0.256 & 1.000 \\
PDF   & 0.655 & 1.000 \\
Trace & \textbf{0.850} & 1.000 \\
\bottomrule
\end{tabular*}
\vspace{-1mm}
\end{table}

As shown in Table~\ref{tab:adjacent-auc}, Trace substantially improves the challenging S1-versus-S2 distinction while retaining perfect separation between S2 and S3.
The highest same-series Trace score is $0.930$, whereas the lowest shared-base score is $0.997$, leaving a clear margin between the two regimes.
Thus, the improved sensitivity to family-level relationships does not compromise shared-base identification.

\section{Intermediate Cross-Series Relationships}
\label{app:cross-series}

The three regimes represent increasing amounts of shared training information, but they do not exhaust all possible lineage relationships.
We therefore examine Qwen2.5 and Qwen3 as an intermediate cross-series case.
These models share a provider and aspects of their design philosophy, while differing in architecture and pretraining corpus.

\begin{table}[!ht]
\centering
\vspace{1mm}
\caption{\small
Trace scores for cross-series Qwen2.5 and Qwen3 model pairs.
Regime-level means are included as reference points.}
\label{tab:cross-series}
\vspace{1mm}
\small
\renewcommand{\arraystretch}{1.08}
\begin{tabular*}{0.58\textwidth}{@{\extracolsep{\fill}}lc}
\toprule
\textbf{Model Pair} & \textbf{Trace} \\
\midrule
Qwen2.5-7B vs.\ Qwen3-14B   & 0.310 \\
Qwen2.5-7B vs.\ Qwen3-8B    & 0.441 \\
Qwen2.5-14B vs.\ Qwen3-8B   & 0.576 \\
Qwen2.5-14B vs.\ Qwen3-14B  & 0.588 \\
\midrule
Cross-series mean           & \textbf{0.479} \\
Independent-origin mean     & 0.311 \\
Same-series mean            & 0.650 \\
\bottomrule
\end{tabular*}
\vspace{-1mm}
\end{table}

Table~\ref{tab:cross-series} shows that the cross-series scores range
from $0.310$ to $0.588$, with a mean of $0.479$.
This mean lies between those of the independent-origin and same-series regimes, while the variation across individual pairs reflects the heterogeneous nature of cross-series relationships.
These results support interpreting Trace as a continuous relatedness signal, with the three regimes serving as reference anchors rather than rigid or exhaustive categories.

\clearpage
\section{Additional Visualizations of Directional Signals}
\label{appen:angle-signal}

\begin{figure}[h]
\vspace{-2mm}
\centering
\setlength{\tabcolsep}{2pt}

\begin{tabular}{ccc}
\multicolumn{3}{c}{\small \textbf{$K$ projection}} \\
\includegraphics[width=0.32\linewidth]{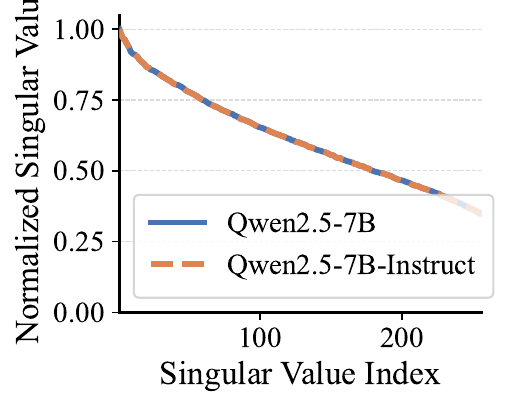} &
\includegraphics[width=0.32\linewidth]{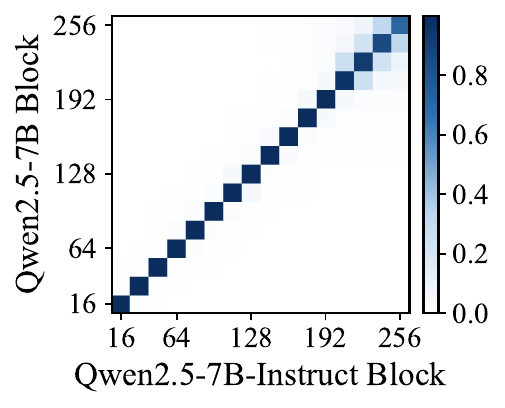} &
\includegraphics[width=0.32\linewidth]{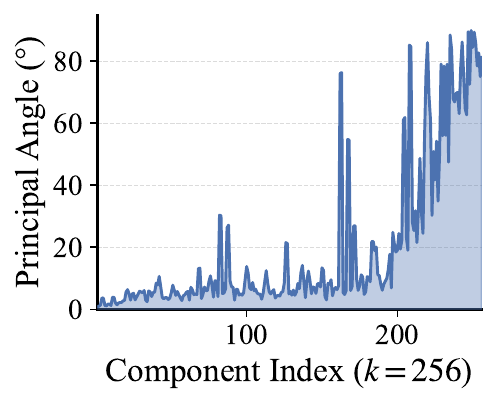} \\
\small (a) Spectral & \small (b) Alignment & \small (c) Angles \\
\end{tabular}

\vspace{2mm}

\begin{tabular}{ccc}
\multicolumn{3}{c}{\small \textbf{$V$ projection}} \\
\includegraphics[width=0.32\linewidth]{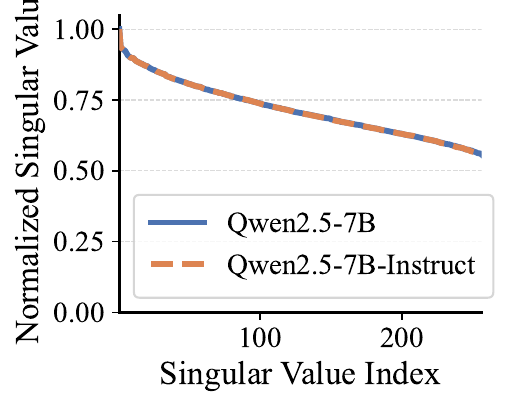} &
\includegraphics[width=0.32\linewidth]{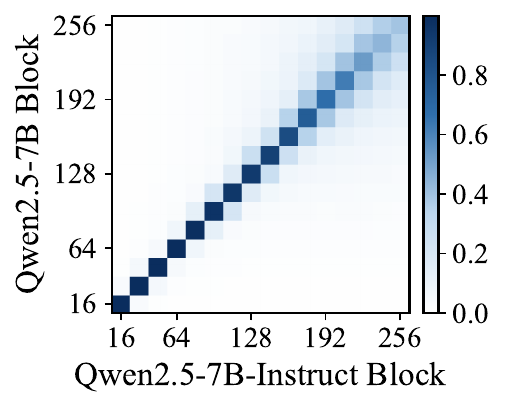} &
\includegraphics[width=0.32\linewidth]{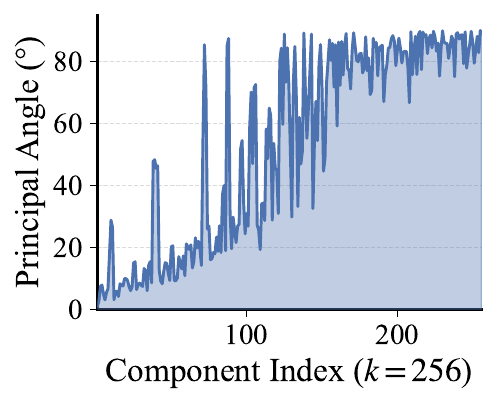} \\
\small (a) Spectral & \small (b) Alignment & \small (c) Angles \\
\end{tabular}

\vspace{2mm}

\begin{tabular}{ccc}
\multicolumn{3}{c}{\small \textbf{$O$ projection}} \\
\includegraphics[width=0.32\linewidth]{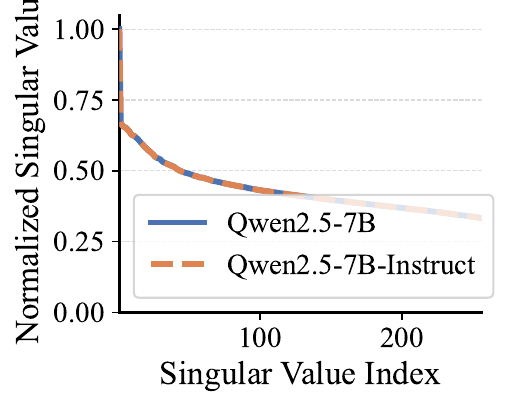} &
\includegraphics[width=0.32\linewidth]{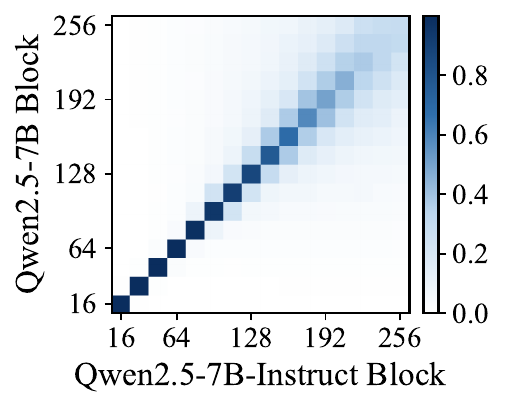} &
\includegraphics[width=0.32\linewidth]{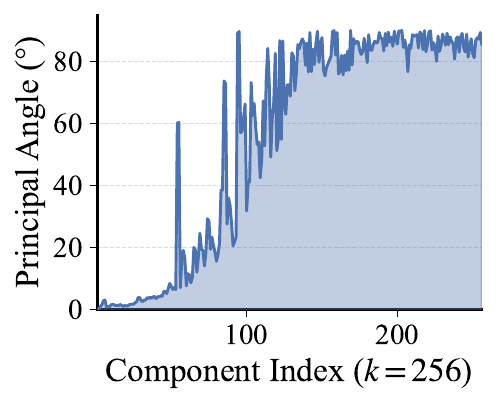} \\
\small (a) Spectral & \small (b) Alignment & \small (c) Angles \\
\end{tabular}

\vspace{2mm}

\begin{tabular}{ccc}
\multicolumn{3}{c}{\small \textbf{$FFN_{\text{GATE}}$ projection}} \\
\includegraphics[width=0.32\linewidth]{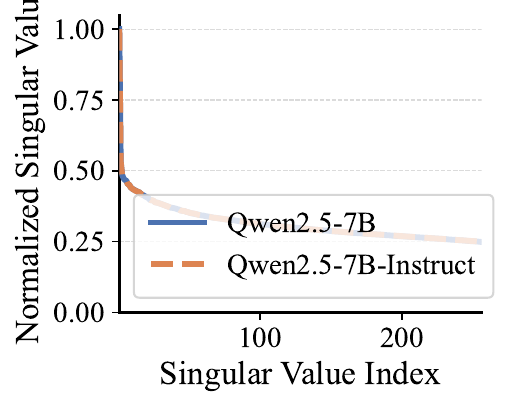} &
\includegraphics[width=0.32\linewidth]{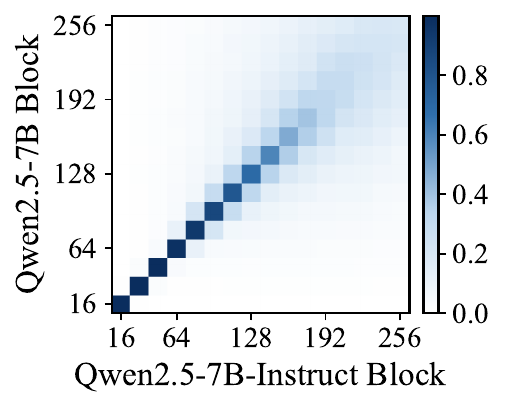} &
\includegraphics[width=0.32\linewidth]{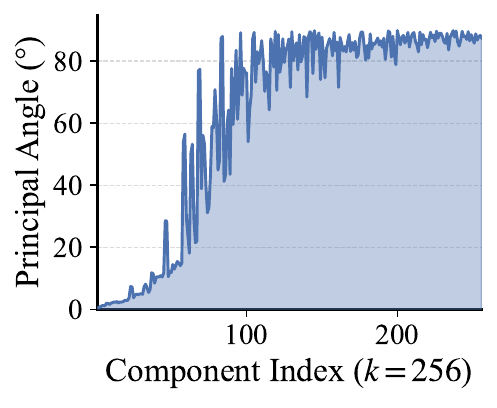} \\
\small (a) Spectral & \small (b) Alignment & \small (c) Angles \\
\end{tabular}

\vspace{-2mm}

\caption{\small
Comparison between a base model (Qwen2.5-7B) and its post-trained variant across multiple feature types ($K$, $V$, $O$, and $FFN_{\text{GATE}}$) at the middle transformer layer, using top-$k=256$ singular components.
Experimental setup keep consistent with Fig\,\ref{fig:rotation-motivation}.
}
\vspace{-2mm}
\label{fig:rotation-multi-feature}
\end{figure}

\vspace{-2mm}
\section{Spectral Directional-based Fingerprint Algorithm}
\vspace{-2mm}
\label{appen:angle-algo}
Alg.~\ref{alg:rotation-fingerprint} summarizes the proposed Subspace Alignment algorithm.
It computes layer-wise similarity from the $J$ smallest singular values of the cross-subspace matrix, then aggregates the least-aligned layers and model components into an overall score.

\vspace{-2mm}
\begin{algorithm}[h]
\caption{Subspace Alignment Fingerprint and Comparison}
\label{alg:rotation-fingerprint}
\small
\begin{algorithmic}[1]

\Require Models $\mathcal{M}_a, \mathcal{M}_b$; components $\mathcal{C}$; top-$k$; $K$ (singular values); $K_{\text{layer}}$
\Ensure Similarity score $S(\mathcal{M}_a, \mathcal{M}_b)$

\For{each component $c \in \mathcal{C}$}

    \Function{ExtractSubspace}{$\mathcal{M}, c$}
        \State $\mathcal{U}_c \gets \emptyset$
        \For{each layer $l$}
            \State $W \gets W_c^{(l)}$
            \State $U \gets \text{Top-}k\text{ left singular vectors of } W$
            \State append $U$ to $\mathcal{U}_c$
        \EndFor
        \State \Return $\mathcal{U}_c$
    \EndFunction

    \State $\mathcal{U}_c^a \gets \textsc{ExtractSubspace}(\mathcal{M}_a, c)$
    \State $\mathcal{U}_c^b \gets \textsc{ExtractSubspace}(\mathcal{M}_b, c)$

    \State $\mathcal{S}_c \gets \emptyset$ \Comment{store layer-wise similarities}

    \For{each layer $l$}
        \State $U_a \gets \mathcal{U}_c^a[l]$, \quad $U_b \gets \mathcal{U}_c^b[l]$
        \State $C \gets U_a^\top U_b$
        \State $\{\sigma_i\} \gets \text{SingularValues}(C)$
        \State $\mathcal{I} \gets$ indices of $K$ smallest $\sigma_i$
        \State $S_c^{(l)} \gets \frac{1}{K} \sum_{i \in \mathcal{I}} \sigma_i$
        \State append $S_c^{(l)}$ to $\mathcal{S}_c$
    \EndFor

    \State $\mathcal{L}_c \gets$ indices of $K_{\text{layer}}$ smallest values in $\mathcal{S}_c$
    \State $S_c \gets \frac{1}{K_{\text{layer}}} \sum_{l \in \mathcal{L}_c} \mathcal{S}_c[l]$

\EndFor
\State $S \gets \frac{1}{|\mathcal{C}|} \sum_{c} S_c$
\State \Return $S$
\end{algorithmic}
\end{algorithm}
\vspace{-2mm}

\vspace{-2mm}
\section{Sensitivity to the Aggregation Parameter $J$}
\vspace{-2mm}
\label{app:j-ablation}

The parameter $J$ in Eq.~\eqref{eq:Sc_subspace} determines how many of the least-aligned singular directions are averaged in the layer-wise Subspace Alignment score.
As shown in Fig.~\ref{fig:rotation-motivation}(c), directional deviations are concentrated in a small subset of the $k=256$ directions.
A smaller $J$ emphasizes the strongest rotations, while a larger $J$ incorporates more well-aligned directions.
Table~\ref{tab:j-ablation} reports results for $J\in\{1,3,5,10,20\}$ across representative shared-base model pairs.
The scores increase monotonically with $J$, while the relative differences across post-training settings remain consistent.
We use $J=3$ to retain sensitivity to localized geometric changes while reducing dependence on a single extreme direction.

\begin{table}[b]
\centering
\vspace{-6mm}
\caption{\small
Subspace Alignment scores under different values of $J$ for representative
shared-base model pairs.
DPO, PPO, and RAFT are compared with Qwen2.5-Math-7B, while Alpaca variants
are compared with LLaMA-3.1-8B.
The main experiments use $J=3$.}
\label{tab:j-ablation}
\vspace{2mm}
\small
\setlength{\tabcolsep}{8pt}
\renewcommand{\arraystretch}{1.08}
\begin{tabular}{lccccc}
\toprule
\textbf{Setting}
& $\boldsymbol{J=1}$
& $\boldsymbol{J=3}$
& $\boldsymbol{J=5}$
& $\boldsymbol{J=10}$
& $\boldsymbol{J=20}$ \\
\midrule
DPO          & 0.875 & \textbf{0.957} & 0.974 & 0.986 & 0.993 \\
PPO          & 0.970 & \textbf{0.975} & 0.993 & 0.996 & 0.998 \\
RAFT         & 0.933 & \textbf{0.961} & 0.986 & 0.992 & 0.996 \\
Alpaca-10\%  & 0.930 & \textbf{0.976} & 0.985 & 0.992 & 0.995 \\
Alpaca-100\% & 0.637 & \textbf{0.889} & 0.921 & 0.959 & 0.979 \\
\bottomrule
\end{tabular}
\end{table}

\section{Additional Experimental Setups of Shared-Base Models}
\label{appen:setup-share-base}

This section documents the controlled experimental procedures for constructing data-scale variants, which enable systematic evaluation of how fine-tuning dataset size affects weight-space geometry.
All shared-base model pairs are cataloged in Appendix\,\ref{appen:model-zoo}.

\textbf{Data-Scale Variants.}
To evaluate sensitivity to dataset scale, we construct controlled variants by fine-tuning the same base model on datasets of different sizes.
We conduct a systematic ablation study using meta-llama/Llama-3.1-8B as the base model, creating four data-scale variants by subsampling the Alpaca instruction dataset~\citep{alpaca}:
\begin{itemize}[leftmargin=*, itemsep=1pt, parsep=0pt]
    \item \textbf{10\% scale}: 5,200 instruction-response pairs (10\% of 52K Alpaca dataset).
    \item \textbf{25\% scale}: 13,000 instruction-response pairs (25\% of dataset).
    \item \textbf{50\% scale}: 26,000 instruction-response pairs (50\% of dataset).
    \item \textbf{100\% scale}: 52,000 instruction-response pairs (full Alpaca dataset).
\end{itemize}

\textbf{Implementation via LLaMA-Factory.}
All data-scale variants are fine-tuned using LLaMA-Factory~\citep{zheng2024llamafactory}, a unified framework for efficient LLM training.
We use the following configuration to ensure perfect experimental control:

\textit{Training hyperparameters}:
\begin{itemize}[leftmargin=*, itemsep=1pt, parsep=0pt]
    \item Learning rate: $1 \times 10^{-5}$ (constant schedule).
    \item Batch size: 128 (across 4 GPUs with gradient accumulation).
    \item Training epochs: 3.
    \item Optimizer: AdamW with $\beta_1=0.9$, $\beta_2=0.999$, weight decay $0.01$.
    \item Gradient clipping: max norm 1.0.
    \item Precision: bfloat16 mixed precision training.
\end{itemize}

\textit{Data preparation}:
For each scale, we perform stratified random sampling from the full Alpaca dataset to ensure balanced representation across instruction categories (e.g., open-ended generation, closed QA, summarization).
The same random seed (42) is used across all scales to ensure that smaller datasets are strict subsets of larger ones (i.e., 10\% $\subset$ 25\% $\subset$ 50\% $\subset$ 100\%).
This nested sampling design enables direct interpretation of scale effects without confounding from data distribution shifts.

\section{Additional Experiments of Fine-Grained Lineage Discrimination}
\label{appen:exp-fine-grain}

\begin{table*}[h]
\centering
\scriptsize
\caption{\small
Similarity scores for Scenario 3 (shared-base) before and after Gaussian perturbation ($\sigma^2=10^{-4}$), including subspace alignment for fine-grained analysis.}
\vspace{2mm}
\setlength{\tabcolsep}{3pt}
\begin{tabular}{lccccc|ccccc}
\toprule
\multirow{2}{*}{\textbf{Model Pair}} 
& \multicolumn{5}{c}{\textbf{Original}} 
& \multicolumn{5}{c}{\textbf{Noise}} \\
\cmidrule(lr){2-6} \cmidrule(lr){7-11}
& AWM & HuReF & PDF & Trace & Subspace
& AWM & HuReF & PDF & Trace & Subspace \\
\midrule

\multicolumn{11}{c}{\textbf{Data-scale variants}} \\
\midrule

Qwen2.5-7B vs Qwen2.5-7B-Instruct & 0.999 & 0.999 & 1.000 & 0.998 & 0.823 & 0.999 & 0.999 & 1.000 & 0.999 & 0.824 \\
Qwen2.5-14B vs Qwen2.5-14B-Instruct & 0.999 & 1.000 & 1.000 & 0.998 & 0.632 & 0.999 & 1.000 & 1.000 & 0.999 & 0.631 \\
Qwen3-4B vs Qwen3-4B-Instruct & 0.998 & 0.999 & 0.999 & 0.999 & 0.208 & 0.999 & 0.999 & 0.999 & 0.999 & 0.209 \\
Qwen3-4B vs Qwen3-4B-Thinking & 1.000 & 0.999 & 0.998 & 0.999 & 0.184 & 1.000 & 0.999 & 0.998 & 0.999 & 0.185 \\

LLaMA-3.1-8B vs LLaMA-3.1-8B-Instruct & 0.998 & 0.997 & 0.999 & 0.997 & 0.814 & 0.999 & 0.998 & 0.999 & 0.998 & 0.810 \\
LLaMA-3.1-8B vs Alpaca-10\% & 0.999 & 0.998 & 1.000 & 1.000 & 0.976 & 0.999 & 0.999 & 1.000 & 1.000 & 0.981 \\
LLaMA-3.1-8B vs Alpaca-25\% & 0.999 & 0.999 & 0.999 & 0.998 & 0.935 & 0.999 & 0.999 & 0.999 & 0.999 & 0.930 \\
LLaMA-3.1-8B vs Alpaca-50\% & 1.000 & 1.000 & 1.000 & 0.998 & 0.903 & 1.000 & 1.000 & 1.000 & 0.999 & 0.909 \\
LLaMA-3.1-8B vs Alpaca-100\% & 1.000 & 1.000 & 1.000 & 0.998 & 0.889 & 1.000 & 1.000 & 1.000 & 0.999 & 0.872 \\

\midrule
\multicolumn{11}{c}{\textbf{Algorithmic variants}} \\
\midrule

LLaMA-3.1-8B vs Tulu-3-8B & 1.000 & 1.000 & 1.000 & 0.999 & 0.787 & 1.000 & 1.000 & 1.000 & 0.999 & 0.789 \\
LLaMA-3.1-8B vs Tulu-3-8B-SFT & 0.999 & 0.999 & 1.000 & 0.998 & 0.783 & 0.999 & 0.999 & 1.000 & 0.999 & 0.783 \\
LLaMA-3.1-8B vs Tulu-3-8B-DPO & 0.998 & 0.999 & 0.999 & 0.999 & 0.783 & 0.999 & 0.999 & 0.999 & 1.000 & 0.780 \\

Qwen2.5-Math-7B vs Qwen2.5-7B-SFT & 1.000 & 1.000 & 0.999 & 0.999 & 0.934 & 1.000 & 1.000 & 0.999 & 0.999 & 0.932 \\
Qwen2.5-Math-7B vs Qwen2.5-7B-DPO & 0.998 & 0.996 & 0.998 & 0.998 & 0.957 & 0.999 & 0.998 & 0.999 & 0.999 & 0.963 \\
Qwen2.5-Math-7B vs Qwen2.5-7B-DPO-Zero & 1.000 & 1.000 & 0.999 & 0.999 & 0.992 & 1.000 & 1.000 & 0.999 & 1.000 & 0.990 \\
Qwen2.5-Math-7B vs Qwen2.5-7B-PPO-Zero & 0.997 & 0.998 & 0.997 & 0.999 & 0.975 & 0.998 & 0.999 & 0.998 & 1.000 & 0.972 \\
Qwen2.5-Math-7B vs Qwen2.5-7B-RAFT-Zero & 1.000 & 1.000 & 0.999 & 0.999 & 0.961 & 1.000 & 1.000 & 0.999 & 1.000 & 0.961 \\

Qwen3-14B vs UniReason-no-think & 0.998 & 0.999 & 0.998 & 0.999 & 0.360 & 0.999 & 0.999 & 0.999 & 0.999 & 0.368 \\
Qwen3-14B vs UniReason-think & 1.000 & 0.999 & 1.000 & 0.999 & 0.358 & 1.000 & 0.999 & 1.000 & 1.000 & 0.352 \\
Qwen3-14B vs UniReason-RL & 0.998 & 0.999 & 0.999 & 0.997 & 0.356 & 0.999 & 0.999 & 0.999 & 0.998 & 0.347 \\

\bottomrule
\end{tabular}
\end{table*}

In the shared-base regime, all magnitude-based metrics (AWM, HuReF, PDF, and Trace) produce near-saturated scores close to 1, reflecting highly aligned global structure inherited from the same pretrained model. 
As a result, they fail to provide meaningful discrimination across different post-training variants.

In contrast, subspace alignment exhibits a significantly wider range of values, revealing clear distinctions across both data-scale and algorithmic variants. 
For example, data-scale variations induce gradual changes in alignment scores, while different optimization strategies (e.g., DPO, PPO, RAFT) lead to more pronounced deviations, indicating stronger geometric transformations in weight space.

Importantly, these patterns remain consistent under Gaussian perturbation, demonstrating that subspace alignment captures robust and intrinsic directional differences. 
These results highlight that fine-grained lineage signals are primarily encoded in subspace geometry rather than spectral magnitude.

\section{Robustness under Model Transformations}
\label{app:transformation-robustness}

We further evaluate the robustness of our framework under four common model transformations in the shared-base setting.
Specifically, we apply INT8 and INT4 quantization to Qwen2.5-7B, unstructured magnitude pruning with sparsity ranging from $10\%$ to $30\%$, SLERP merging between the DPO and PPO-Zero variants of Qwen2.5-Math-7B, and data distillation by comparing DeepSeek-R1-Distill-Qwen-7B with its Qwen2.5-Math-7B base model.
As shown in Table~\ref{tab:transformation-robustness}, Trace remains between $0.999$ and $1.000$ across all evaluated transformations, indicating that the coarse-grained shared-base lineage signal is consistently preserved.
Subspace Alignment provides a complementary measure of transformation severity: it remains high under INT8 quantization and model merging, but decreases under INT4 quantization, stronger pruning, and data distillation.
Across these settings, the K and V projections remain largely unchanged, whereas Q, O, and FFN components exhibit substantially larger directional deviations.

\begin{table}[h]
\centering
\vspace{2mm}
\caption{\small
Robustness under quantization, unstructured magnitude pruning, model merging,
and data distillation.
Trace and overall Subspace Alignment are reported together with component-wise
Subspace Alignment scores.}
\label{tab:transformation-robustness}
\vspace{2mm}

\footnotesize
\setlength{\tabcolsep}{3.1pt}
\renewcommand{\arraystretch}{1.10}

\begin{tabular}{llccccccccc}
\toprule
\textbf{Transformation}
& \textbf{Setting}
& \textbf{Trace}
& \textbf{Sub.}
& \textbf{K}
& \textbf{V}
& \textbf{Q}
& \textbf{O}
& \textbf{Gate}
& \textbf{Up}
& \textbf{Down} \\
\midrule

Quantization
& INT8
& 1.000 & 0.947 & 1.000 & 1.000 & 0.933 & 0.938 & 0.937 & 0.915 & 0.906 \\

Quantization
& INT4
& 1.000 & 0.357 & 1.000 & 1.000 & 0.150 & 0.098 & 0.115 & 0.090 & 0.048 \\

\midrule

Pruning
& 10\%
& 1.000 & 0.855 & 1.000 & 1.000 & 0.888 & 0.777 & 0.784 & 0.758 & 0.777 \\

Pruning
& 20\%
& 1.000 & 0.679 & 1.000 & 1.000 & 0.686 & 0.635 & 0.489 & 0.471 & 0.474 \\

Pruning
& 30\%
& 1.000 & 0.491 & 1.000 & 1.000 & 0.462 & 0.305 & 0.241 & 0.224 & 0.203 \\

\midrule

SLERP merge
& vs.\ Base
& 1.000 & 0.965 & 1.000 & 1.000 & 0.977 & 0.986 & 0.934 & 0.983 & 0.875 \\

SLERP merge
& vs.\ DPO
& 1.000 & 0.978 & 1.000 & 1.000 & 0.991 & 0.995 & 0.923 & 0.984 & 0.951 \\

SLERP merge
& vs.\ PPO-Zero
& 1.000 & 0.972 & 1.000 & 1.000 & 0.985 & 0.996 & 0.925 & 0.987 & 0.910 \\

\midrule

Distillation
& vs.\ Base
& 0.999 & 0.401 & 1.000 & 1.000 & 0.112 & 0.339 & 0.111 & 0.112 & 0.129 \\

\bottomrule
\end{tabular}

\vspace{2mm}
\end{table}

\end{document}